\documentclass[10pt,twocolumn,letterpaper]{article}

\usepackage[pagenumbers]{cvpr} 

\usepackage[dvipsnames]{xcolor}

\usepackage{enumitem}
\usepackage{bm}
\usepackage{url}
\usepackage{comment}
\usepackage{graphicx}
\usepackage{colortbl}
\usepackage{xcolor}
\usepackage{amsmath}
\usepackage{multirow}
\usepackage{adjustbox}
\usepackage{threeparttable}
\usepackage{epsfig}
\usepackage{wrapfig}
\usepackage{algorithm}
\usepackage{algorithmic}
\usepackage{setspace}
\usepackage{bm}
\usepackage{float}
\usepackage{makecell}

\usepackage{booktabs}       
\usepackage{amsfonts}       
\usepackage{nicefrac}       
\usepackage{microtype}      
\usepackage{tablefootnote}
\def\pz{{\phantom{0}}}
\definecolor{cvprblue}{rgb}{0.21,0.49,0.74}
\usepackage[pagebackref,breaklinks,colorlinks,allcolors=cvprblue]{hyperref}
\def\paperID{****} 
\def\confName{CVPR}
\def\confYear{2026}
\title{Guiding a Diffusion Transformer with the Internal Dynamics of Itself}

\author{\hspace{-0.5cm}
Xingyu Zhou$^1$\quad Qifan Li$^1$\quad
Xiaobin Hu$^2$\quad Hai Chen$^3$$^4$\quad Shuhang Gu$^1$\thanks{Corresponding author.  \hfill \textbf{Project page:} {\href{https://zhouxingyu13.github.io/Internal-Guidance/}{Internal Guidance}}}\\
\hspace{-0.5cm}
$^1$University of Electronic Science and Technology of China \hspace{0pt}
$^2$National University of Singapore \hspace{0pt}\\
$^3$Sun Yat-sen University \hspace{0pt}
$^4$North China Institute of Computer Systems Engineering\\
\hspace{-0.5cm}
{\tt\small \{xy.chous526, shuhanggu\}@gmail.com}}

\begin{document}
\maketitle
\begin{figure*}[t]
\centering
\includegraphics[width=1\textwidth]{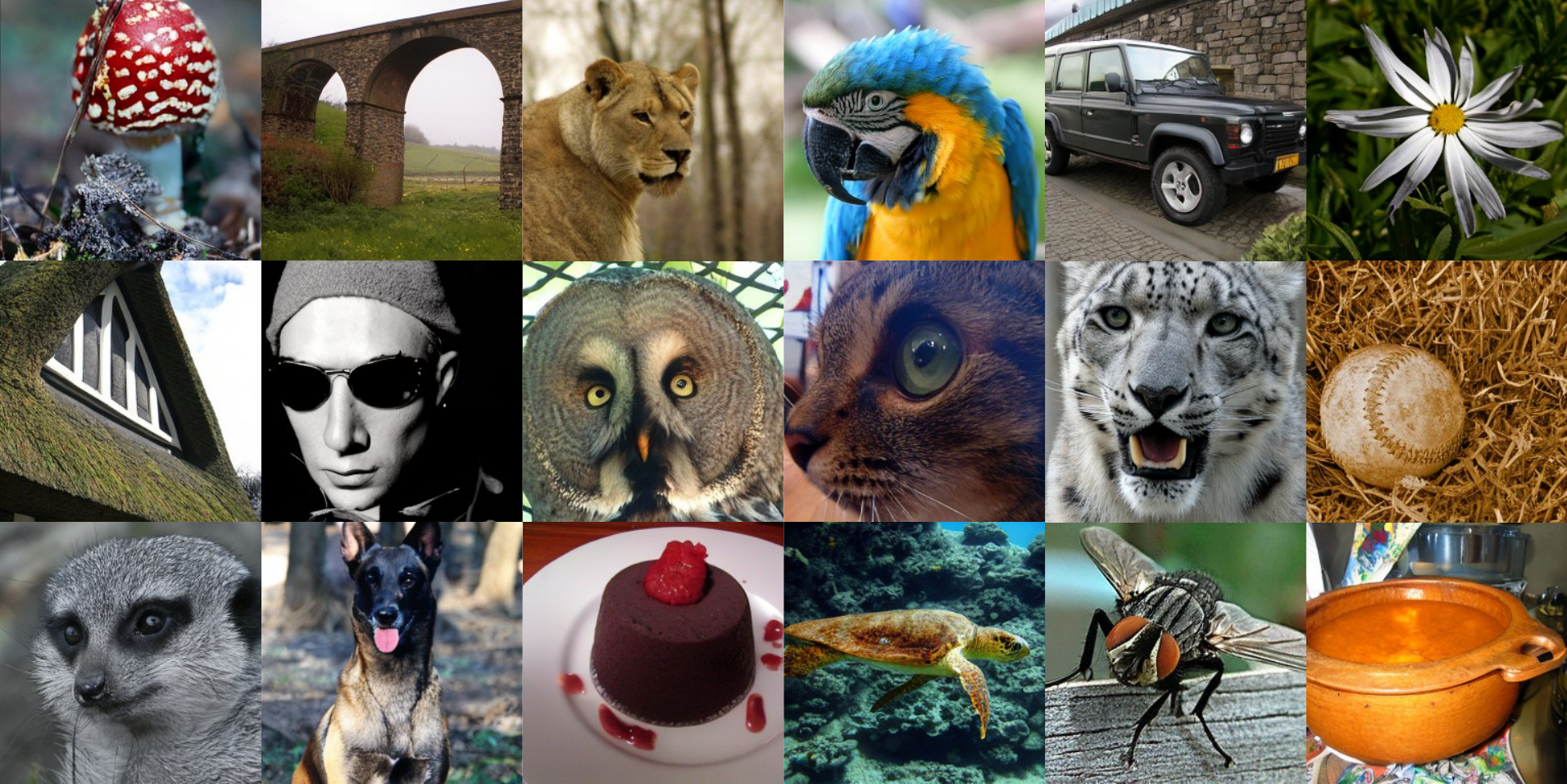}
\caption{\textbf{Visualization Results.} We visualize our latent diffusion system with proposed IG together with LightningDiT-XL trained on ImageNet 256 × 256 resolution. With an IG scale of 1.4 and a CFG scale of 1.45, and further combining the guidance interval, we ultimately achieved the state-of-the-art FID = 1.19.  More uncurated samples are provided in \textbf{Supplementary Material}.}
\vspace{-0.2cm}
\label{fig:visual}
\end{figure*}
\begin{abstract}
The diffusion model presents a powerful ability to capture the entire (conditional) data distribution.
However, due to the lack of sufficient training and data to learn to cover low-probability areas, the model will be penalized for failing to generate
high-quality images corresponding to these areas.
To achieve better generation quality, guidance strategies such as classifier free guidance (CFG) can guide the samples to the high-probability areas during the sampling stage.
However, the standard CFG often leads to over-simplified or distorted samples. 
On the other hand, the alternative line of guiding diffusion model with its bad version is limited by carefully designed degradation strategies, extra training and additional sampling steps. 
In this paper, we proposed a simple yet effective strategy \textbf{\textit{Internal Guidance (IG)}}, which introduces an auxiliary supervision on the intermediate layer during training process and extrapolates the intermediate and deep layer's outputs to obtain generative results during sampling process.
This simple strategy yields significant improvements in both training efficiency and generation quality on various baselines. 
On ImageNet 256×256, SiT-XL/2+IG achieves FID=5.31 and FID=1.75 at 80 and 800 epochs. 
More impressively, LightningDiT-XL/1+IG achieves FID=1.34 which achieves a large margin between all of these methods.
Combined with CFG, LightningDiT-XL/1+IG achieves the current state-of-the-art FID of 1.19. 
\end{abstract}    
\section{Introduction}
\label{sec:intro}
Denoising Diffusion models have achieved notable success in high-dimensional data generation and rapidly applied to fields such as images, videos and 3D. 
Recently, the introduction of the Transformer architecture has further enhanced the scalability of the diffusion model in these fields \cite{peebles2023scalable, ma2024sit, esser2024scaling}, demonstrating its outstanding performance.
As the scale of the diffusion model continues to increase, how to efficiently achieve excellent generation results has become a highly focused research topic.

The training objective of the diffusion model is to cover the entire (conditional) data distribution. 
The model is severely penalized for failing to cover low-probability areas, but it lacks sufficient training and data to learn to generate high-quality images corresponding to these areas.
Several studies introduce additional guidance strategies during the sampling stage to 
encourage the generative process to focus on high-probability areas of the data distribution.
Among them, classifier-free guidance (CFG) \cite{ho2022classifier} leads to a better alignment between the prompt and the result by avoiding the unconditional results.
Nevertheless, an excessively high CFG coefficient may push the sampling trajectory beyond the expected range of the conditional distribution, resulting in over-simplified or distorted samples \cite{kynkaanniemi2024applying}.
An alternative line of research attempts to guide a well-trained model using a bad version of itself  \cite{karras2024guiding} instead of relying on CFG.
By avoiding the results of the corresponding degraded version of the generative model during the sampling stage, the generation quality is improved while avoiding overly simplified results.
Although promising, these approaches still require carefully designed degradation strategies \cite{hong2023improving, hong2024smoothed}, extra training \cite{karras2024guiding} or additional sampling steps \cite{hyung2025spatiotemporal, chen2025s}, thereby limiting their potential for the large-scale application.

To address this, we propose a simple yet effective sampling guidance strategy, named \textbf{\textit{Internal Guidance (IG)}}, which utilizes the intermediate-layer outputs within the Diffusion Transformer to enhance generation quality.
Specifically, we introduce an auxiliary supervisory signal at an intermediate layer during training process, enabling the model to produce a weaker generative output from mid-level representations.
Building upon this, during the sampling stage, we leverages the relationship between intermediate and deep outputs during sampling to extrapolate.
This internal guidance achieves an Autoguidance-like \cite{karras2024guiding} effect that maintains diversity and improves the generation quality but without any additional sampling steps.
Moreover, we show that IG can also combine with the guidance interval \cite{kynkaanniemi2024applying} or CFG to further enhance generative performance.
Besides providing intermediate results for guidance, we observe that the introduction of intermediate supervision alone can alleviate partial vanishing gradients. It's convergence performance comparable to, or even better than, those achieved with self-supervised regularization applied to intermediate layers.
Further leveraging the insight in IG, we discuss how modeling the discrepancy between intermediate and deep outputs within the loss function can inspire the design of new training acceleration strategies.

We conduct comprehensive experiments to evaluate the effect of our proposed internal guidance. 
Our IG brings significant performance improvements to both DiTs, SiTs and LightningDiT.
Ultimately, we compared our generative performance with the current state-of-the-art models on SiT-XL/2 and LightningDiT-XL/1 with IG.
For SiT-XL/2 training, we show the model achieves FID = 5.31 and FID = 1.75 on class-conditional ImageNet generation only using 80 and 800 training epochs (without CFG) which already exceeds the FID of the vanilla SiT-XL at 1400 epoch and REPA at 800 epochs that uses pre-trained representation model to align.
Moreover, with classifier-free guidance and guidance interval, our scheme shows an improved FID at 1.46 at 800 epochs.
And for LightningDiT-XL/1 model which uses self-supervised representation-based tokenizer, it achieves FID = 2.42 using 60 epochs and further achieves FID = 1.34 at 680 epochs, which achieves a large margin between all of these methods.
Combined with classifier-free guidance and guidance interval, our scheme achieves state-of-the-art results of FID = 1.19 at 680 epochs.

\section{Related Work}
\label{Related Work}
We discuss with the most relevant literature here and provide a more discussion in \textbf{Supplementary Material}.

\noindent\textbf{Sampling Guidance in Diffusion Models.}
The sampling guidance after the training has become the key to improving the quality of the visual generation.
Classifier-free guidance (CFG) \cite{ho2022classifier} has become the standard method which forms a guided score by extrapolating from an unconditional one.
While highly effective, CFG will cause the problem of reduced generation diversity when its guiding coefficient becomes larger.
Several studies \cite{chang2023muse, gao2023masked, kynkaanniemi2024applying, sadat2023cads, wang2024analysis} have reduced the downsides of CFG by making the guidance weight noise level-dependent.
Another research direction separates the class guidance part from the CFG and only uses a weaker version of the current model for guidance.
This approach improves image quality while maintaining diversity. 
While promising, these methods are limited by the complex design of degradation strategies \cite{hong2023improving, hong2024smoothed}, extra training \cite{karras2024guiding} or additional sampling steps \cite{hyung2025spatiotemporal, chen2025s}, which are not available for current large-scale image generators in practice.
In this paper, our proposed internal guidance can be seamlessly integrated into the current advanced diffusion Transformer with almost no additional cost, demonstrating its potential for large-scale application.

\noindent\textbf{Intermediate Representation Regularization for Diffusion Transformers.}
Many recent works have attempted to introduce intermediate representation regularization in diffusion transformer training.
MaskDiT \citep{zheng2023fast} and SD-DiT \cite{zhu2024sd} combine MAE's \cite{he2022masked} and IBOT's \cite{zhou2021ibot} training strategy into the DiT's training process. 
TREAD \citep{krause2025tread} designs a token routing strategy to speed up Diffusion Transformer’s training.
More recently, self-supervised learning has been introduced into diffusion training to further enhance training efficiency.
These methods can be broadly categorized into: (1) aligning intermediate network representations with pre-trained self-supervised models \cite{yu2024representation, leng2025repa, wu2025representation}, and (2) incorporating self-supervised representation learning as a regularization mechanism during training \cite{jiang2025sra, wang2025diffuse}.
In this paper, we employ the simplest approach of introducing an auxiliary supervision loss to alleviate the gradient vanishing problem in deep networks.
This design bears conceptual similarity to DeepFlow \cite{shin2025deeply}, but it mainly focuses on introducing second-order ODEs in the training of the diffusion Transformer.
Interestingly, 
we observed that the convergence performance of this simple method is comparable to that achieved by incorporating self-supervised learning as a form of regularization.
\section{Methodology}
\label{sec:Method}
\subsection{Preliminaries}
Our work is based on Flow-matching \cite{lipman2022flow} and Diffusion \cite{ho2020denoising}, \citep{ma2024sit} and \cite{albergo2023stochastic} have proposed a unified perspective to understand these two types of generative modeling methods. 
We first introduce the relevant preliminaries.
The standard Flow-matching and Diffusion-based models both start from Gaussian noise $\epsilon \sim \mathcal{N}(0, I)$ and gradually transform it into data samples $\mathbf{x}_* \sim p(\mathbf{x})$ through stochastic processes.
This process can be expressed as
\begin{equation}
    \mathbf{x}_t = \alpha_t \mathbf{x}_* + \sigma_t \epsilon,
\end{equation}
where $\alpha_t$ and $\sigma_t$ are decreasing and increasing functions of time $t \in [0, T]$. 
The difference between the generative models based on Flow-matching and Diffusion lies in that the former typically interpolates noise and data within a finite time interval $t \in [0, 1]$, while the latter defines a forward stochastic differential equation (SDE), which converges to a Gaussian distribution $\mathcal{N}(0, I)$ as $t \to \infty$. The denoising network $ D_\theta$ estimates the original data by minimizing the denoising diffusion objective:
\begin{equation}
    \mathbb{E}_{\mathbf{x}_* \sim p_{\mathbf{x}}} \mathbb{E}_{\epsilon \sim \mathcal{N}(0, \mathbf{I})} \| D_\theta(\mathbf{x}_t, t) - \mathbf{x}_0 \|^2 .
\end{equation}

As the network size of generative models continues to increase, how to efficiently obtain high-quality generation results has become the focus of attention. Next, we will introduce our framework from two aspects: the loss function and the corresponding sampling guidance after training completion. 
Furthermore, we will discuss some related properties of the proposed sampling guidance, as well as the design of new training acceleration methods based on this.
\begin{figure}[t]
\centering
\includegraphics[width=0.46\textwidth]{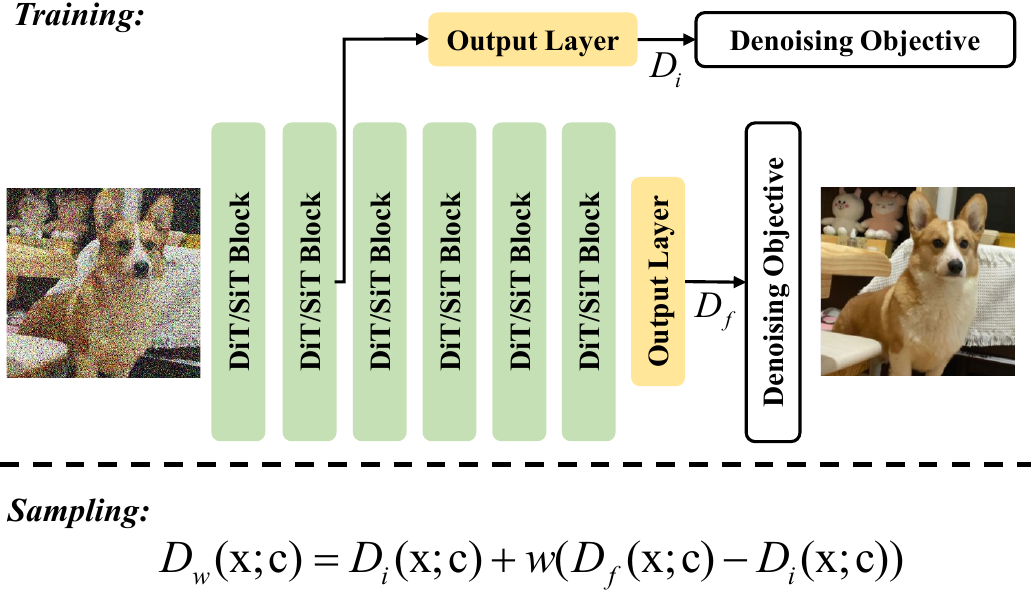}
\caption{\textbf{The overall framework of our proposed Internal Guidance.} We introduce an additional auxiliary supervision loss during training, and utilize the intermediate layer outputs during sampling process to guide the final outputs.}
\vspace{-0.4cm}
\label{fig:framework}
\end{figure}

\subsection{Internal Guidance: Guiding a Diffusion Model with Its Own Internal Dynamics}

\noindent\textbf{Training a Diffusion Model with Intermediate Supervision.}
The difficulty of training deep networks has been explored in many aspects in various fields \cite{wang2015training, ioffe2015batch, he2016deep, he2015delving}.
Recently, in the field of generation, people have introduced self-supervised representation learning as a regularization method for training diffusion models \cite{jiang2025sra,wang2025diffuse}.
In this paper, we attempt the simplest solution, which is to apply an auxiliary supervision in the intermediate layers of the network.
Specifically, we additionally define an output layer after the intermediate part of the network.
Accordingly, the intermediate and final losses $\mathcal{L}_{\textit{inter}}$ and $\mathcal{L}_{\textit{final}}$ are defined as
\begin{equation}
    \begin{cases}
    \mathcal{L}_{\textit{inter}} = \| D_i(\mathbf{x}_t, t) - \mathbf{x}_0 \|^2 ,\\
    \mathcal{L}_{\textit{final}} = \| D_f(\mathbf{x}_t, t) - \mathbf{x}_0 \|^2 ,
    \end{cases}  
\end{equation}
where $D_i(\cdot)$ and $D_f(\cdot)$ denote the outputs of the intermediate and final layers, respectively. 
The overall framework is illustrated in Figure~\ref{fig:framework}.
For instance, for the training of a deep Diffusion Transformer, its final training loss becomes:
\begin{equation}
    \mathcal{L} = \mathcal{L}_{\textit{final}} + \lambda\mathcal{L}_{\textit{inter}},
\end{equation}
where $\lambda > 0$ is a hyperparameter that controls the trade-off between the loss of the intermediate output and that of the final output.
As shown in Table \ref{tab:supervise}, We verified the effect of this loss on the SiT-B/2 model \cite{ma2024sit}.
Although this method is very simple, it can surprisingly achieve results that are similar to or even better than those obtained by using the regularization method based on representation learning for complex designs \cite{jiang2025sra, wang2025diffuse}.

\noindent\textbf{Sampling a Diffusion Model with Its Internal Outputs.}
The sampling stage is crucial for obtaining high-quality generated results.
The standard classifier-free guidance (CFG) \cite{karras2024guiding} sampling strategy can effectively improve the generation quality, but will cause an undesirable reduction in diversity. 
\begin{figure}[!t]
\centering
\includegraphics[width=0.48\textwidth]{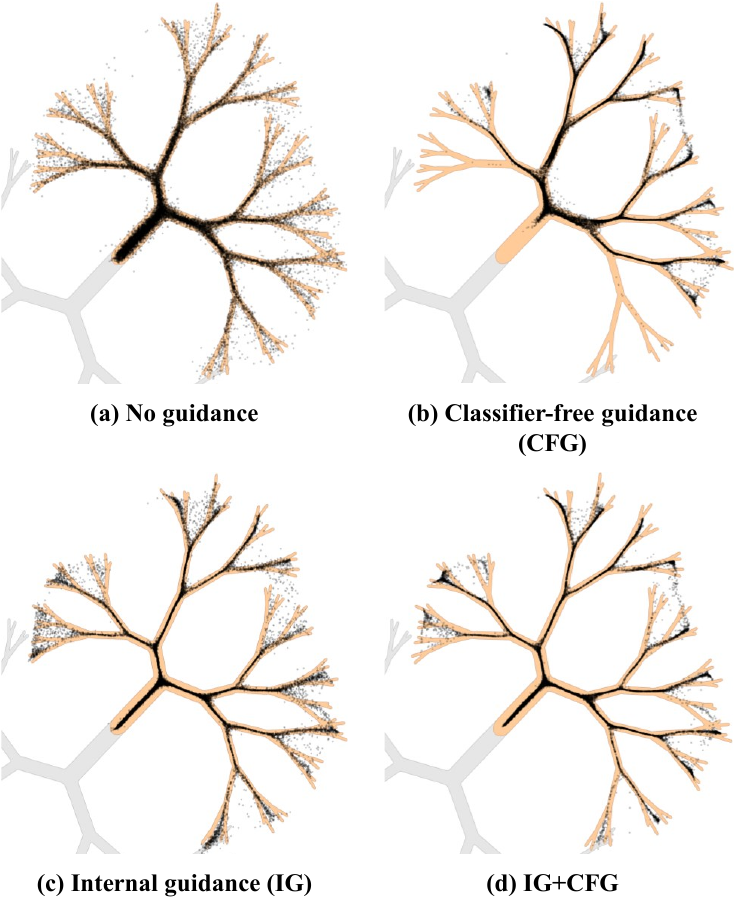}
\caption{\textbf{A fractal-like 2D distribution with two classes indicated with gray and orange regions.}
Approximately 99\% of the probability mass is inside the shown contours. (a) Conditional sampling using a small denoising
diffusion model generates outliers due to its limited fitting capability. (b) Classifier-free guidance (w = 2.5) eliminates the vast majority of outliers but reduces diversity by over-emphasizing the class.
(c) Internal guidance (w = 2) can maintain diversity as Autoguidance \cite{karras2024guiding} while allowing some outliers at the ends of the branches. (d) The combination of IG and CFG can significantly reduce outliers without reducing diversity.}\label{fig:igcfg}
\vspace{-0.4cm}
\end{figure}
An alternative line of research attempts to maintain diversity by guiding a well-trained model using a bad version of itself \cite{karras2024guiding}.
While promising, these approaches are still limited by carefully designed degradation strategies \cite{hong2023improving, hong2024smoothed}, extra training \cite{karras2024guiding} or additional sampling steps \cite{hyung2025spatiotemporal, chen2025s}.
By applying an auxiliary supervision to the intermediate layer of the denoising model in the training stage,
we can obtain the output of the final layer and that of its intermediate layer during the sampling stage.
This naturally amounts to creating a bad version corresponding to the current final layer's output.
In this way, we propose \textit{\textbf{Internal Guidance (IG)}}, the intermediate output $D_i$ can be used to guide the final output $D_f$ to shift away from its own poor distribution, thereby achieving better generation results.
Similar to classifier free guidance (CFG), its guiding effect is achieved by extrapolating the two denoised results using a factor $w$
\begin{equation}\label{eq:ig}
    D_w(\mathbf{x};\mathbf{c}) = D_i(\mathbf{x};\mathbf{c}) + w (D_f(\mathbf{x};\mathbf{c})-D_i(\mathbf{x};\mathbf{c})).
\end{equation}

Unlike the original Autoguidance \cite{karras2024guiding} strategy, our proposed IG is a plug and play approach that no longer requires specific training on particular bad versions for specific experimental setups or additional sampling consumption.
And this sampling method can also enhance the quality of image generation while maintaining diversity.
Building upon this, we conduct an extensive discussion on the properties of IG.
\section{Discussion}\label{discuss}
In this section, we conduct a detailed analysis of the specific impacts of IG from two perspectives: (1) \textit{the compatibility between IG and CFG} and (2) \textit{Guidance Interval in IG}.

\subsection{The compatibility between IG and CFG}
Although Internal Guidance (IG) can significantly improve the generation results without additional consumption during the sampling stage, 
its combination with Classifier-Free Guidance (CFG) can further improve the generation quality.
To further elucidate the compatibility between our proposed IG and CFG, we conducted a series of visualization experiments to analyze at which levels these guidance strategies improve generative quality.
We study a 2D toy example like \cite{karras2024guiding} where a small-scale denoising
network is trained to perform conditional diffusion in a synthetic dataset (Figure \ref{fig:igcfg}). 
The dataset is designed to exhibit low local dimensionality (i.e., highly anisotropic and narrow support) and a hierarchical emergence of local details upon noise removal. 
Both properties that can be expected from the actual manifold of realistic images \cite{brown2022verifying, pope2021intrinsic}. Details of the experimental setup are provided in the \textbf{Supplementary Material}.

As shown in Figure \ref{fig:igcfg}(a), sampling from a diffusion model without guidance produces many outliers outside the data distribution.
When using CFG, this guidance pulls samples toward higher values of $\log\left[ p_{con}(\mathbf{x}\mid \mathbf{c})/p_{uncon}(\mathbf{x}\mid \mathbf{c}) \right]$ \cite{karras2024guiding}, where $p_{con}$ and $p_{uncon}$ are the conditional and unconditional distributions learned by the denoising diffusion model.
As shown in Figure \ref{fig:igcfg}(b), increasing the CFG coefficient shifts samples away from the opposite class and reduces outliers, but also suppresses branches near other classes, which enhances inter-class separability while reducing diversity.
In contrast, Autoguidance \cite{karras2024guiding} provides class-independent gradients that point inward towards the data manifold.
The trajectories of these samples generated by this well trained model generally follow the local orientation and branching of the data manifold, pushing the samples deeper into the “good side” concentrates them at the manifold. 

Furthermore, our proposed IG exhibits a performance similar to that of Autoguidance (Figure \ref{fig:igcfg}(c)).
When the IG coefficient increases, these samples produce a narrower distribution than the ground truth, but in practice this does not appear to have an adverse effect on the images.
However, this reduced intra-class sample distance may introduce new outliers when a large Autoguidance or IG coefficient is applied before the model is fully trained. In this case, the stronger output fails to accurately point toward high-probability regions, rendering the guidance less effective.
Because CFG provides class-conditional directionality, it complements IG by further suppressing such outliers. Combining the two achieves superior generative quality across training stages (Figure \ref{fig:igcfg}(d)).
In Section \ref{ablation}, we present a detailed quantitative validation to demonstrate this effect.
\subsection{Guidance Interval in Internal Guidance}
Meanwhile, the guidance interval \cite{kynkaanniemi2024applying} can also affect the performance of internal guidance. 
The original guidance interval encourages using large CFG coefficient for sampling within a continuous interval of noise levels, which is helpful for achieving better generation performance. 
For our proposed internal guidance, we also redefined the guidance function of Equation \ref{eq:ig} by replacing $\omega$ with a piecewise constant function:
\begin{equation}
\begin{aligned}
&    D_w(\mathbf{x};\mathbf{c}) = D_i(\mathbf{x};\mathbf{c}) + w(\sigma) (D_f(\mathbf{x};\mathbf{c})-D_i(\mathbf{x};\mathbf{c})), \\
&    \text{where } w(\sigma) = \begin{cases} w & \text{if } \sigma \in (\sigma_{low}, \sigma_{high}], \\ 1 & \text{otherwise,} \end{cases}
\end{aligned}
\end{equation}
where, $\sigma$ is the coefficient that controls the intensity of the noise.
In the original guidance interval, applying CFG during periods of high noise range drastically reduces the variation in the results, basically leading them towards a handful of “template images” per class.
Therefore, the guidance interval used in the CFG indicates that this guidance should only be used in the middle and low-noise ranges.
For our proposed IG with the guidance interval, we observed the opposite phenomenon: the internal guidance does not need to be applied in the low-noise range, but applied in the high-noise and middle-noise ranges.
In Section \ref{ablation}, we have provided a detailed presentation of the generated results after applying the guidance interval to the IG sampling.

\section{Experiments}\label{sec:experiments}
\subsection{Setup}
\begin{table}[t]
\centering\small
\captionof{table}{
\textbf{Auxiliary supervision loss on different intermediate layers} on ImageNet 256$\times$256 with internal guidance ($w=1.5$). $\downarrow$ and $\uparrow$ indicate whether lower or higher values are better.}\label{tab:interloss}
\resizebox{0.4\textwidth}{!}{
\begin{tabular}{c c  c c c}
\toprule
\multirow{2}{*}{Block Layers}  &\multicolumn{2}{c}{Baseline}&\multicolumn{2}{c}{+IG}\\
\cmidrule(lr){2-3} \cmidrule(lr){4-5} 
& {FID$\downarrow$} & {IS$\uparrow$}& {FID$\downarrow$} & {IS$\uparrow$}\\
\midrule
\rowcolor{gray!20}
SiT-B/2~\citep{ma2024sit} & 33.02 & 43.71& -- & --\\
\midrule
\cellcolor{yellow!10}2  & \textbf{30.45} & \textbf{47.97} & 20.39 & 61.38\\
\cellcolor{yellow!10}4  & 30.60 & 47.70 & \textbf{19.02} & \textbf{65.06}\\
\cellcolor{yellow!10}6  & 34.15 & 42.38 &25.04 & 52.39\\
\cellcolor{yellow!10}8  & 38.05 & 37.97 &34.05 & 40.35\\
\cellcolor{yellow!10}10 &36.37&40.91&35.58&40.86\\
\cellcolor{yellow!10}2 and 6  & 33.34 & 43.08&31.55&46.66\\
\bottomrule
\end{tabular}}
\vspace{-0.2em}
\end{table}
\begin{table}[t]
\centering\small
\vspace{-1.0em}
\vspace{0.2em}
\captionof{table}{
\textbf{Convergence analysis of the SiT-B/2 model} on ImageNet 256$\times$256 without internal guidance and classifier-free guidance (CFG). $\downarrow$ and $\uparrow$ indicate whether lower or higher values are better, respectively. All models were trained for 80 epochs.
}\label{tab:supervise}
\resizebox{0.46\textwidth}{!}{
\begin{tabular}{c c c c}
\toprule
Methods & Learning Paradigms & {FID$\downarrow$} & {IS$\uparrow$} \\
 
\midrule
\rowcolor{gray!20}
\multicolumn{2}{l}{SiT-B/2 Baseline~\citep{ma2024sit}} & 33.02 & 43.71 \\

\midrule
SRA~\citep{jiang2025sra} & Self-Supervised & {29.10} & {50.20} \\
Disperse Loss~\citep{wang2025diffuse} & Self-Supervised & {31.45} & {47.05} \\
\midrule
\rowcolor{yellow!10}
Auxiliary Supervision & Supervised & {30.45} & {47.97} \\
\bottomrule
\end{tabular}}
\vspace{-0.8em}
\end{table}
\noindent\textbf{Implementation details.} 
For experiments on DiT~\citep{peebles2023scalable} and SiT~\citep{ma2024sit}, we strictly follow their original setup. 
For experiments on  LightningDiT~\citep{yao2025reconstruction}, we find using the recipe in LightningDiT leads to instability or abnormal training dynamics at early epochs and slow EMA model convergence at early epochs. 
To solve these issues, we instead use the Muon optimizer \cite{jordan2024muon} instead of AdamW and change the EMA weight from 0.9999 to 0.9995.
Other optimization hyperparameters are the same as vanilla LightningDiT.
We use ImageNet-1K \cite{deng2009imagenet} for diffusion training, where each image is preprocessed to the resolution of 256×256 and follow ADM for other data preprocessing protocols. 
Each image is then encoded into a compressed vector $\mathbf{z} \in \mathbb{R}^{32\times32\times4}$ using the Stable Diffusion VAE or $\mathbf{z} \in \mathbb{R}^{16\times16\times32}$ using the VA-VAE.
Detailed experimental details and hyperparameter settings, are provided in the \textbf{Supplementary Material}.

\noindent\textbf{Evaluation protocol.} We report Fr\'echet inception distance (FID) \cite{heusel2017gans}, sFID \cite{nash2021generating}, inception score \cite{salimans2016improved}, precision(Pre.) and recall(Rec.) \cite{kynkaanniemi2019improved} using 50000 samples. 
Specifically, we uniformly randomly sample the 1,000 class labels 50,000 times and generate images accordingly SD, SiT, REPA for a fair comparison.
Sampling follows original DiT, SiT and LightningDiT that we use the SDE Euler-Maruyama sampler with 250 steps for the experiments on DiT and SiT, and use the ODE Heun sampler with 125 steps for the experiments on LightningDiT.
Full evaluation protocol details are also provided in the \textbf{Supplementary Material}.

\subsection{Ablation Studies}\label{ablation}

\noindent\colorbox{yellow!10}{\textbf{The position of intermediate supervision.}} 
We begin by examining the effect of attaching the auxiliary supervision loss to different layers, which are shown in Table \ref{tab:interloss}.
We find that regularizing only the first few layers (e.g., 4 in SiT-B/2 \citep{ma2024sit} at 80 epochs) in training is effective.
However, placing the auxiliary supervision loss in the latter half of the network or adding multiple auxiliary supervision losses in multiple intermediate layers cannot improve the convergence speed.
We hypothesize that it interferes with the training of the deep layer's output.
Furthermore, we are surprised to find that the convergence effect with the auxiliary supervision loss is comparable to that of methods incorporate self-supervised representation learning regularization (Table \ref{tab:supervise}).
In large-scale experiments, we apply the auxiliary supervision loss and internal guidance to the 8-th layer of the large-scale diffusion transformer model.

\begin{figure}[t]
\centering
\includegraphics[width=0.38\textwidth]{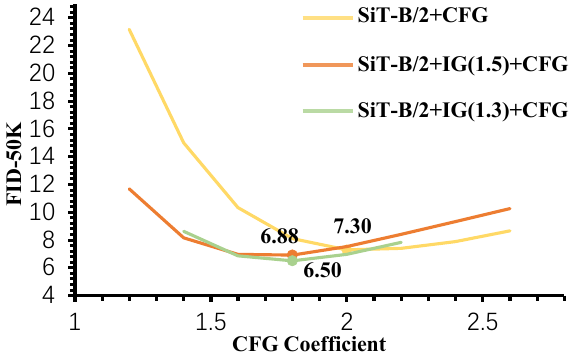}
\caption{\colorbox{green!5}{\textbf{The combination of IG and CFG}} can yield a higher FID value, which is superior to simply apply CFG. In particular, the FID value generated by the combination of a lower IG coefficient and CFG is higher than that of a higher coefficient.}
\vspace{-0.2cm}
\label{fig:cfgscale}
\end{figure}
\begin{table}[t]
\centering\small
\vspace{-0.4em}
\captionof{table}{\textbf{The IG coefficient and the corresponding guidance interval analysis} on ImageNet 256$\times$256. $\downarrow$ and $\uparrow$ indicate whether lower or higher values are better, respectively.
} \label{tab:interval}
\vspace{-0.3em}
\begin{tabular}{c c c c  c c c c c}
\toprule
Sampling & $\omega$ & Guidance Interval & {FID$\downarrow$}& {IS$\uparrow$}\\
 
\midrule
\rowcolor{gray!20}
w/o IG & 1.0 & $[0, 1)$& 30.60 & 47.70 \\

\midrule
\multirow{7}{*}{w/ IG} & \cellcolor{green!5} 1.3 & $[0, 1)$ & 21.88 & 59.55\\
& \cellcolor{green!5} 1.5 & $[0, 1)$ & 19.02 & 65.06\\
& \cellcolor{green!5} 1.7 & $[0, 1)$ & 17.61 & 68.16\\
 & \cellcolor{green!5} 1.9 &$[0, 1)$ & \textbf{17.38} & \textbf{69.12}\\
 & \cellcolor{green!5} 2.1 & $[0, 1)$ & 17.75 & 68.67 \\
& \cellcolor{green!5} 2.3 &$[0, 1)$ & 18.53 & 66.77 \\
 & 2.3 & \cellcolor{cyan!5}$[0.3, 1)$& \textbf{16.19} & \textbf{72.95} \\
 & 2.3 & \cellcolor{cyan!5}$[0.3, 0.7)$& 18.23 & 66.21  \\
 & 2.3 & \cellcolor{cyan!5}$[0, 0.7)$ & 20.69 & 60.14 \\
\bottomrule
\end{tabular}
\vspace{-0.6cm}
\end{table}
\noindent\colorbox{green!5}{\textbf{Compatibility between IG and CFG.}}
We also demonstrate that the proposed IG combined with CFG can further enhance the generation quality in the quantitative experiment. 
We apply IG on the fourth layer of SiT-B/2 to conduct the subsequent experiments.
We conducted a rough search for the optimal combination rules of IG and CFG.
As shown in Figure \ref{fig:cfgscale}, after SiT-B/2 combines with our IG and then applies CFG, the best generated metric FID can be reduced from the original optimal 7.30 with only CFG to 6.50. 
Meanwhile, we find that when combining CFG, using a lower IG coefficient can yield better generation results, which is different from the optimal coefficient when only adding IG alone (Table \ref{tab:interval} and Figure \ref{fig:ablation}).
And this process basically does not involve significant computational costs, which indicates that the IG can serve as an easily integrable component for modern generative models to enhance the generation quality.

\begin{figure}[t]
\centering
\vspace{-0.1cm}
\includegraphics[width=0.42\textwidth]{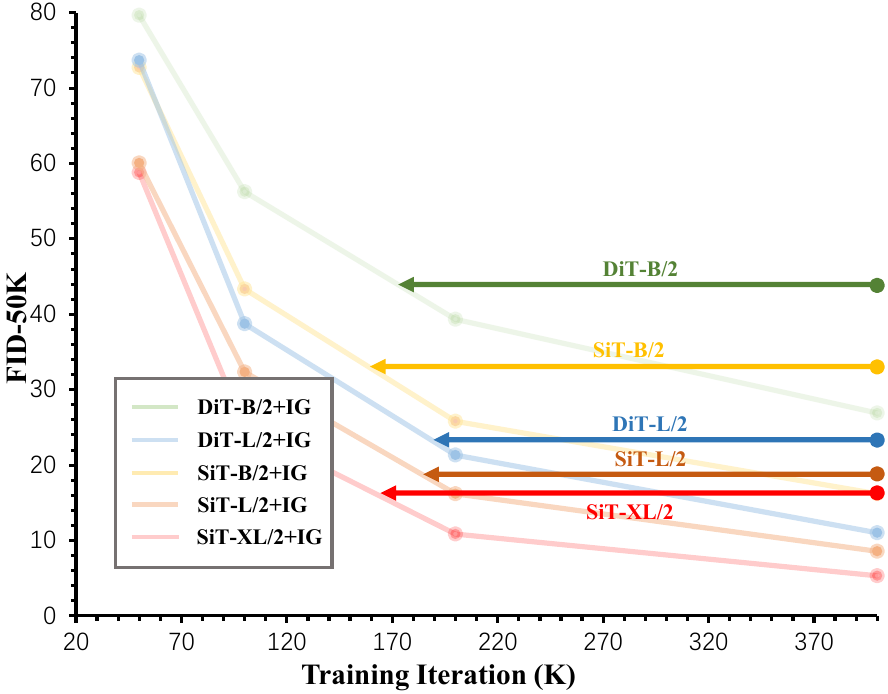}
\caption{\textbf{\colorbox{orange!5}{Scalability of IG.}} The relative improvement of IG over the vanilla model (DiTs and SiTs) becomes increasingly significant as the model size grows.}
\vspace{-0.1cm}
\label{fig:scaling}
\end{figure}

\noindent\colorbox{cyan!5}{\textbf{Guidance interval in internal guidance.}} 
Next, we examine the role of guidance interval in internal guidance.
%
We also apply IG on the fourth layer of SiT-B/2 to conduct experiments.
In the Table \ref{tab:interval}, we can see that when no guidance interval is used, the optimal FID is approximately around the IG coefficient $\omega$ of 1.9.
After increasing the IG coefficient, we find it actually has a positive effect on the generation quality when the low-noise range does not apply IG.
This is somewhat contrary to the conclusion of guidance interval used in CFG \cite{kynkaanniemi2024applying}.
In our rough search, the best generative results are achieved when the IG coefficient is 2.3 and the guidance interval range is $[0.3, 1)$.

\begin{table}[t]
\centering\small
\vspace{-0.8em}
\captionof{table}{
\textbf{Ablation study for $\lambda$.}
}\label{tab:lambda}
\vspace{-0.6em}
\begin{tabular}{c c c c c}
\toprule
$\lambda$ & \cellcolor{red!5}{0.25} & \cellcolor{red!5}{0.5} & \cellcolor{red!5}{0.75} & \cellcolor{red!5}{1} \\
 
\midrule

{FID$\downarrow$} & 30.38 & {30.60} & 31.58 & 31.24 \\
{IS$\uparrow$} & 47.86 & {47.70} & 46.23  & 46.24\\
\bottomrule
\end{tabular}
\vspace{-1.9em}
\end{table}
\noindent\colorbox{orange!5}{\textbf{Scalability.}} We investigate the scalability of the auxiliary supervision loss with internal guidance by varying the model sizes of the diffusion transformers. 
In general, the effect of achieving efficient generation results of IG becomes more significant as the diffusion transformer model increases in size.
We demonstrate this by plotting FID-50K of different SiT and DiT models with IG and without CFG in Figure: IG achieves the same FID level more quickly with larger models.
To facilitate comparison, we set the IG coefficient to 2.3 and the guidance interval range is $[0.3, 1)$ in all the DiTs and SiTs models.
And the IG is applied in the fourth layer of the B/2 scale model and in the eighth layer of other scales for the experiments.

\noindent\colorbox{red!5}{\textbf{Effect of $\lambda$.}} Finally, we examine the effect of the regularization coefficient $\lambda$ by training SiT-B/2 models for 400K with different coefficients 0.25 to 1.0 and comparing the performance.
As shown in Table \ref{tab:lambda}, the performance becomes stable after $\lambda<= 0.5$.

\subsection{Comparsion with State-of-the-art Methods}

\begin{table*}[h]
\centering
\small
\renewcommand{\arraystretch}{0.9}
\setlength{\tabcolsep}{8pt} 
\caption{\textbf{Class-conditional performance on ImageNet 256$\times$256.} SiT + IG reaches an FID of 1.75 in and further achieves an FID of 1.46 with CFG. LightningDiT + IG reaches an FID of 1.34 without CFG, outperforming all prior methods by a large margin, and achieves an FID of 1.19 with CFG. For a fair comparison, all prior methods employ \textbf{random sampling} 50K samples across 1,000 class labels.
}
\label{tab:comparison_perf}
\makebox[\textwidth][c]{
\resizebox{1\textwidth}{!}{
\begin{tabular}{l c c ccccc ccccc}
\toprule
\multirow{2}{*}{\textbf{Method}} & \multirow{2}{*}{\makecell{\textbf{Epochs}}} & \multirow{2}{*}{\textbf{\#Params}} & \multicolumn{5}{c}{\textbf{Generation@256 w/o CFG}} & \multicolumn{5}{c}{\textbf{Generation@256 w/ CFG or Autoguidance}} \\
\cmidrule(lr){4-8} \cmidrule(lr){9-13}
 & & & \textbf{FID}$\downarrow$& \textbf{sFID}$\downarrow$ & \textbf{IS}$\uparrow$ & \textbf{Prec.}$\uparrow$ & \textbf{Rec.}$\uparrow$ & \textbf{FID}$\downarrow$& \textbf{sFID}$\downarrow$ & \textbf{IS}$\uparrow$ & \textbf{Prec.}$\uparrow$ & \textbf{Rec.}$\uparrow$ \\
\arrayrulecolor{black}\midrule
\multicolumn{11}{l}{\textit{\textbf{Pixel Diffusion}}} \\
\arrayrulecolor{black!30}\midrule
ADM~\citep{dhariwal2021diffusion} &  400  &  554M  & 10.94& 6.02 &  101.0 & 0.69 & 0.63 & 3.94& 6.14 & 215.8 & \textbf{0.83} & 0.53\\
RIN~\citep{jabri2022scalable} &  480  &  410M  & 3.42& -  & 182.0  &  -   & -&  -   &  -   &   -    &  -   &  -  \\
PixelFlow~\citep{chen2025pixelflow} & 320 & 677M & - & - & -  &   -  &  -   & 1.98& 5.83 & 282.1 & 0.81 & 0.60 \\
PixNerd~\citep{wang2025pixnerd} & 160 & 700M & -   &  - & -      &  - &  -  &  2.15& 4.55 & 297.0 & 0.79 & 0.59 \\
SiD2~\citep{hoogeboom2024simpler} &  1280   &   - & -  & -   &  - &  -   &  -   &  1.38   &   - & -   &  -   &  - \\
\arrayrulecolor{black}\midrule
\multicolumn{11}{l}{\textit{\textbf{Vanilla Latent Diffusion}}} \\
\arrayrulecolor{black!30}\midrule
DiT~\citep{peebles2023scalable} & 1400 & 675M         & 9.62& 6.85 & 121.5 & 0.67 & 0.67 & 2.27& 4.60 & 278.2 & \textbf{0.83} & 0.57 \\
MaskDiT~\citep{zheng2023fast} & 1600 & 675M & 5.69 & 10.34& 177.9 & 0.74 & 0.60 & 2.28& 5.67 & 276.6 & 0.80 & 0.61 \\
SiT~\citep{ma2024sit} & 1400 & 675M         & 8.61& 6.32 & 131.7 & 0.68 & 0.67 & 2.06& 4.50 & 270.3 & 0.82 & 0.59 \\
TREAD~\citep{krause2025tread} & 740 & 675M & -& - & - & - & - & 1.69 & 4.73& 292.7 & 0.81 & 0.63 \\
MDTv2~\citep{gao2023mdtv2} & 1080 & 675M & -& - & - & - & - & 1.58 & 4.52& 314.7 & 0.79 & 0.65 \\
 \midrule
\multirow{2}{*}{\textbf{SiT+IG (ours)}} & 80 & \multirow{2}{*}{678M} & 5.31& 5.68 & 147.7 & \textbf{0.80} & 0.56 & -- & --& -- & -- & -- \\
& 800 &  & 1.75& 4.98 & 228.6 & \textbf{0.80} & 0.62 &1.46 & 4.79& 265.7 & 0.80 & 0.64\\
\arrayrulecolor{black}\midrule
\multicolumn{11}{l}{\textit{\textbf{Latent Diffusion with Self-supervised Representation Model}}} \\
\arrayrulecolor{black!30}
\midrule
\multirow{2}{*}{REPA~\citep{yu2024representation}} & 80 & \multirow{2}{*}{675M} & 7.90 & 5.06& 122.6 & 0.70 & 0.65 & -& - & - & - & - \\
 & 800 &  & 5.90& 5.69 & 157.8 & 0.70 & \textbf{0.69} & 1.42& 4.70 & 305.7 & 0.80 & 0.65 \\
\midrule
\multirow{2}{*}{REPA-E~\citep{leng2025repa}} & 80 & \multirow{2}{*}{675M} & 3.46& 4.17 & 159.8 & 0.77 & 0.63 & 1.67& 4.12 & 266.3 & 0.80 & 0.63 \\
 & 800 &  & 1.83& - & 217.3 & - & - & 1.26& 4.11 & 314.9 & 0.79 & 0.66 \\
 \midrule
\multirow{2}{*}{REG~\citep{wu2025representation}} & 80 & \multirow{2}{*}{677M} & 3.40& - & 184.1 & - & - & 1.86 & 4.49& \textbf{321.4} & 0.76 & 0.63 \\
& 480 &  & 2.20& 4.67 & 219.1 & 0.77 & 0.66 & 1.40& 4.24 & 296.9 & 0.77 & 0.66\\
\midrule
\multirow{2}{*}{LightningDiT~\citep{yao2025reconstruction}} & 64 & \multirow{2}{*}{675M} & 5.14 & 4.22& 130.2 & 0.76 & 0.62 & 2.11& 4.16 & 252.3 & 0.81 & 0.58  \\
 & 800 &  & 2.17& 4.36 & 205.6 & 0.77 & 0.65 & 1.35& 4.15 & 295.3 & 0.79 & 0.65 \\
\midrule
RAE (DiT$^{\text{DH}})$~\citep{zheng2025diffusion} & 800 & 839M & 1.60& 5.40 & \textbf{242.7} & 0.79 & 0.65 & 1.28& 4.72 & 262.9 & 0.78 & \textbf{0.67}\\
\midrule
\multirow{2}{*}{\textbf{LightningDiT+IG (ours)}} & 60 & \multirow{2}{*}{678M} & 2.42& \textbf{3.81} & 173.7 & 0.79 & 0.62 & -- & -- & -- & --& -- \\
 & 680 & & \textbf{1.34}& 3.94 & 229.3 & 0.78 & 0.65 & \textbf{1.19} & \textbf{4.11}& 269.0 & 0.79 & 0.66 \\
\arrayrulecolor{black}\bottomrule
\end{tabular}
}
}
\vspace{-0.2cm}
\end{table*}

We use several recent diffusion-based generation methods as baselines, each employing
different inputs and network architectures.
Specifically, we consider the following three types of approaches:
(a) \textit{Pixel Diffusion}: ADM~\citep{dhariwal2021diffusion}, RIN~\citep{jabri2022scalable}, PixelFlow~\citep{chen2025pixelflow}, PixNerd~\citep{wang2025pixnerd} and SiD2~\citep{hoogeboom2024simpler}, (2) \textit{Vanilla Latent Diffusion}: DiT~\citep{peebles2023scalable}, MaskDiT~\citep{zheng2023fast}, SiT~\citep{ma2024sit}, TREAD~\citep{krause2025tread}, MDTv2~\citep{gao2023mdtv2}, (c) \textit{Latent Diffusion with Self-supervised Representation Model}: REPA~\citep{yu2024representation}, REPA-E~\citep{leng2025repa} and REG~\citep{wu2025representation}, LightningDiT~\citep{yao2025reconstruction} and DiT$^{\text{DH}}$~\citep{zheng2025diffusion}. 
We initialized our models on SiT and LightningDiT respectively to separately verify the effectiveness of the IG in the latent diffusion model with original VAE and aligned VAE that represents the latent space using the pretrained self-supervised representation model (i.e., DINOv2-B \cite{oquab2023dinov2}).

We compare the FID values between SiT-XL/2 and LightningDiT-XL/1 with IG with other state-of-the-art methods without CFG.
As shown in Table \ref{tab:comparison_perf}, IG shows consistent and significant improvement across all model variants under random sampling. We provide balanced sampling results in the \textbf{Supplementary Material}.
In particular, on SiT-XL/2, using IG (1.8) leads to FID=5.31 at 80 epochs, which already exceeds the FID of the vanilla SiT-XL at 1400 epoch and REPA at 800 epochs which uses pre-trained representation model to align.
Note that the performance continues to improve with longer training; for instance, with SiT-XL/2, FID becomes 1.75 at 800 epochs. 
On LightningDiT-XL/1, using IG (1.7) leads to FID=2.42 at 60 epochs, which also already exceeds the FID of the vanilla LightningDiT-XL/1 at 64 epochs.
For longer training, the FID becomes 1.34 at 680 epochs with LightningDiT-XL/1, which achieve a large margin between all of these methods.

Finally, we provide a quantitative comparison between SiT-XL/2 or LightningDiT-XL/1 with IG and other recent diffusion model methods using CFG.
At 800 epochs, SiT-XL/2 + IG (1.8) achieves FID of 1.46 with a CFG scale of $w=1.35$ and an extra guidance interval.
Meanwhile, LightningDiT/1 + IG (1.4) achieves state-of-the-art FID of 1.19 with a CFG scale of $w=1.45$  and an extra guidance interval at 680 epochs.
We provide visual results of LightningDiT-XL/2 + IG in Figure \ref{fig:visual}.
Moreover, we also provide more uncurated examples and experimental results on ImageNet 512×512 in the \textbf{Supplementary Material}, we show that IG also provides significant improvements in such setup.

\begin{figure*}[h]
\centering
\includegraphics[width=1\textwidth]{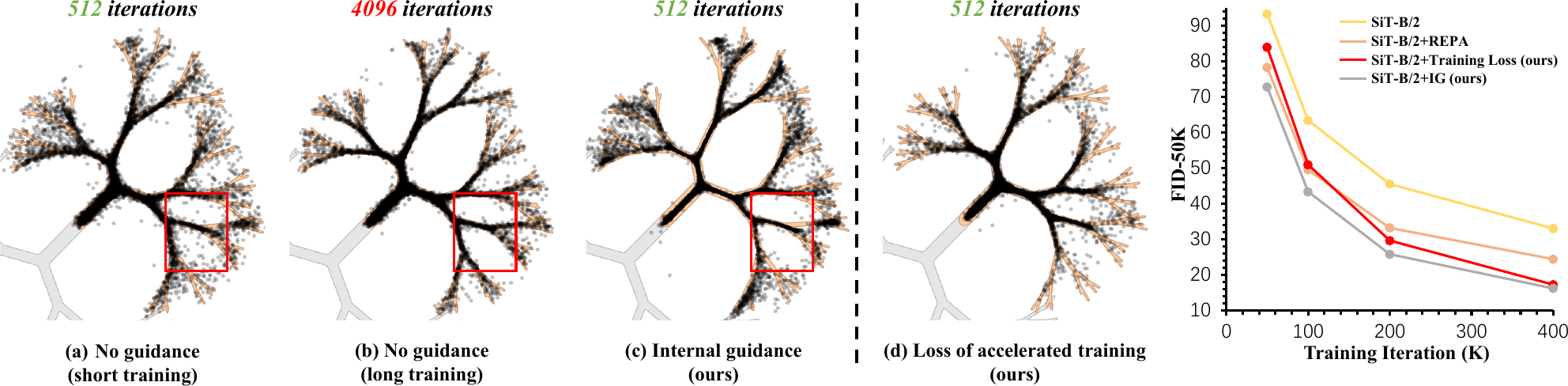}
\caption{\textbf{Internal guidance inspires new training acceleration methods.} (a) Conditional sampling using a not well-trained denoising diffusion model generates a large number of outliers. (b) Outliers can gradually be eliminated with sufficient training. (c) Internal guidance can also eliminate outliers with not well-trained  denoising diffusion model. (d) The loss function we proposed can accelerate convergence. In SiT-B/2 experiments, our proposed loss function demonstrates superior accelerated convergence performance compared to REPA \cite{yu2024representation}.}
\vspace{-0.4cm}
\label{fig:trainprocess}
\end{figure*}

\section{Further Discussion on Training Acceleration}
After conducting the above analysis, we naturally came up with a new question: \textit{\textbf{Can the effect of this IG be replicated during the training process, so as to achieve the goal of training acceleration?}}
We also demonstrated the denoising diffusion model training process through a 2D toy example.
As shown in Figure \ref{fig:trainprocess}(a) and (b), comprehensive training can significantly reduce many outliers near the branches compared to incomplete training.
With our proposed IG, since its guiding direction points towards the overall direction of the stronger model, outliers near the branches can also be effectively reduced (Figure \ref{fig:trainprocess}c).
Based on the above observations, we incorporate the guiding effect of IG into the training process of the denoising diffusion model.
Specifically, we integrate the gradient vectors obtained by calculating the differences between the final and intermediate outputs during the sampling process into the loss function, enabling the model to learn the guidance signal during training.
Similar to \cite{tang2025diffusion}, we rewrite the training objectives for the intermediate and final layers during the training process
\begin{equation}
    \begin{cases}
        \text{Intermediate objective: } \mathbf{x}_0, \\
        \text{Final objective: } \mathbf{x}_0 + \textcolor{purple}{\omega \cdot \text{sg}(D_f(\mathbf{x}_t) - D_i(\mathbf{x}_t))}.
    \end{cases}
\end{equation}
Where $\omega > 0$ is a hyperparameter that controls the strength of the guiding signal for the final layer's training objective.
In practice, we utilize the EMA model that updates along with the model to obtain the outputs at both the intermediate and final layers during the training process and set $\omega = 0.5$ following \cite{tang2025diffusion}.
As shown in Figure \ref{fig:trainprocess}(d), the outliers of all branches have been significantly reduced after modifying the training objective.
In the subsequent SiT-B/2 experiment, our designed training loss significantly outperformed the REPA \cite{yu2024representation} using pre-trained self-supervised representation models after 400K iterations, demonstrating performance comparable to that of IG.
Due to the flexibility and tunability of IG in generating images during the reasoning process, 
we suggest continuing to use IG in actual applications.
\begin{figure}[h]
\centering
\includegraphics[width=0.47\textwidth]{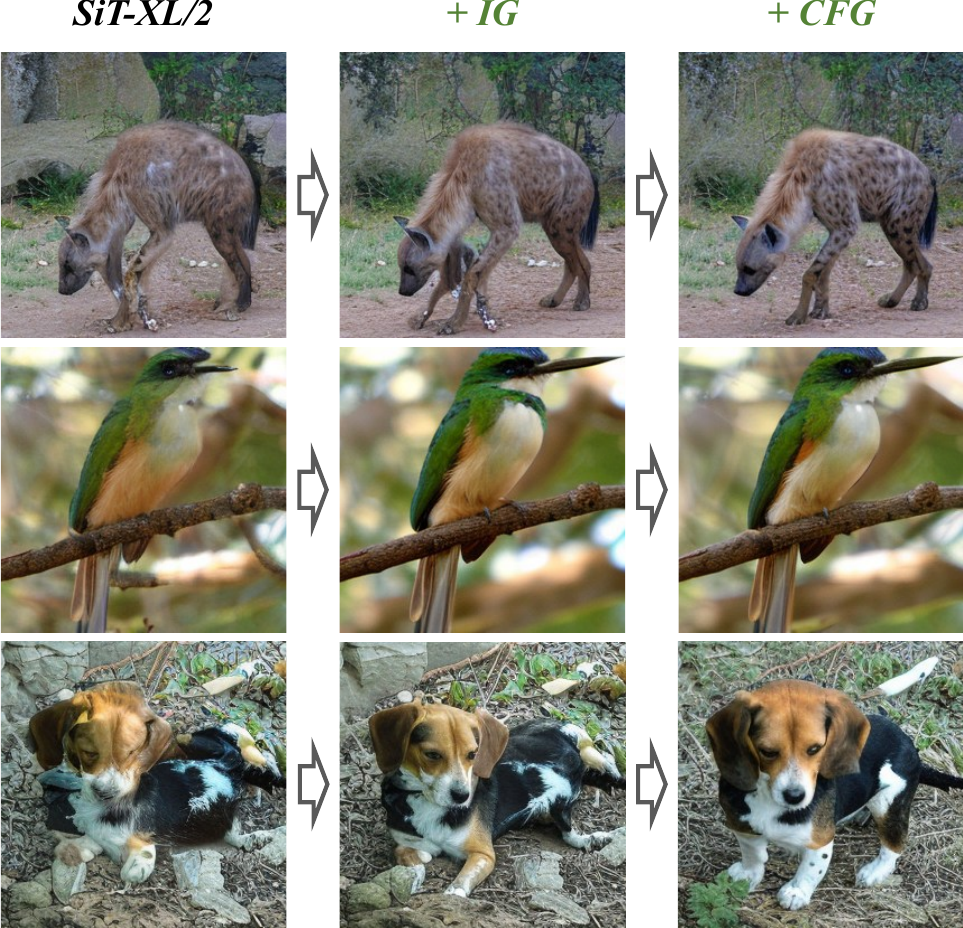}
\vspace{-0.1cm}
\caption{\colorbox{green!5}{\textbf{The combination of IG and CFG}} can produce better visual quality. When only adding IG, better generation results can be obtained. Incorporating CFG can further reduce the inaccurate content in the generated results.}
\vspace{-0.7cm}
\label{fig:ablation}
\end{figure}
\section{Conclusion}
In this paper, we have presented \textbf{\textit{Internal Guidance (IG)}}, a simple but effective guidance strategy for enhancing both generation quality and training efficiency.
We introduce an auxiliary supervision loss in the deep Diffusion Transformer and utilize the output of its intermediate layers to guide the output results at the deep level during the sampling stage. 
This plug-and-play method has significantly improved the quality of image generation.
We further analyze the properties of each part of our proposed IG, including its compatibility with CFG, its own guidance interval and discuss the design of new training acceleration methods based on its insight.
Our IG can significantly improve the generation performance of various scale diffusion transformers with less computational resources.
Ultimately, we get the state-of-the-art FID = 1.19 result for image generation when using LightningDiT-XL/1+IG with additional CFG.
We hope our work would facilitate the research on more applicable sampling methods for visual generation.
\clearpage
\noindent\textbf{Acknowledgement}.
This work was supported by National Natural Science Foundation of China (No. 62476051). 

{
    \small
    \bibliographystyle{ieeenat_fullname}
    \bibliography{main}
}

\clearpage
\setcounter{page}{1}
\maketitlesupplementary
\appendix
In this file, we provide more implementation and experimental details which are not included in the main text. 
In Section \ref{f}, we provide the details of all the evaluation metrics and further compare the generation performance under uniform sampling.
In Section \ref{a}, we provide a detailed comparison of computational resource consumption during the sampling stage between the proposed IG and REPA.
In Section \ref{b}, we provide a comparison of the generate results under different sampling guidance methods.
In Section \ref{c}, we provide the experimental results of the proposed IG on the 512 $\times$ 512 ImageNet dataset.
In Section \ref{d}, we provide more hyperparameter settings and implementation details for Diffusion Transformer experiments of various scales.
In Section \ref{e}, We provide hyperparameter settings and implementation details for the 2D toy example.
In Section \ref{g}, We provide a more detailed discussion on the related work of our approach.
In Section \ref{h}, we provide more uncurated generation results on ImageNet 256×256 from the SiT-XL+IG with CFG.
\begin{table*}[h]
\centering
\small
\renewcommand{\arraystretch}{0.9}
\setlength{\tabcolsep}{8pt} 
\caption{\textbf{Class-conditional performance on ImageNet 256$\times$256 with balanced sampling.} LightningDiT + IG reaches an FID of 1.24 without CFG, outperforming all prior state-of-the-art VQ-based and Diffusion-based methods by a large margin, and achieves an FID of 1.07 with CFG. For a fair comparison, all prior methods employ \textbf{uniform balanced sampling} 50K samples across 1,000 class labels.
}
\label{tab:bc}
\makebox[\textwidth][c]{
\resizebox{1\textwidth}{!}{
\begin{tabular}{l c c cccc cccc}
\toprule
\multirow{2}{*}{\textbf{Method}} & \multirow{2}{*}{\makecell{\textbf{Epochs}}} & \multirow{2}{*}{\textbf{\#Params}} & \multicolumn{4}{c}{\textbf{w/o CFG}} & \multicolumn{4}{c}{\textbf{w/ CFG or Autoguidance}} \\
\cmidrule(lr){4-7} \cmidrule(lr){8-11}
 & & & \textbf{FID}$\downarrow$ & \textbf{IS}$\uparrow$ & \textbf{Prec.}$\uparrow$ & \textbf{Rec.}$\uparrow$ & \textbf{FID}$\downarrow$& \textbf{IS}$\uparrow$ & \textbf{Prec.}$\uparrow$ & \textbf{Rec.}$\uparrow$ \\
\midrule
\multicolumn{11}{l}{\textit{\textbf{Autoregressive}}} \\
\arrayrulecolor{black!30}\midrule
VAR~\citep{tian2024visual} &  350  &  2.0B  & 1.92 & \textbf{323.1}  & \textbf{0.82} & 0.59 & 1.73 & \textbf{350.2} & 0.82 & 0.60\\
MAR~\citep{li2024autoregressive} &  800  &  943M  & 2.35 &  227.8 & 0.79 & 0.62 & 1.55 & 303.7 & 0.81 & 0.62\\
xAR~\citep{ren2025beyond} &  800  & 1.1B  & - & - & - & - & 1.24 & 301.6 & \textbf{0.83} & 0.64\\
\arrayrulecolor{black}\midrule
\multicolumn{11}{l}{\textit{\textbf{Latent Diffusion with Self-supervised Representation Model}}} \\
\arrayrulecolor{black!30}
\midrule
REPA~\citep{yu2024representation}& 800 & 675M & 5.78 & 158.3 & 0.70 & 0.68 & 1.29 & 306.3 & 0.79 & 0.64 \\
DDT~\citep{wang2025ddt}& 400 & 675M & 6.27 & 154.7 & 0.68 & \textbf{0.69} & 1.26 & 310.6 & 0.79 & 0.65\\
%
REPA-E~\citep{leng2025repa}& 800 & 675M & 1.70& 217.3 & 0.77 & 0.66 & 1.15 & 304.0 & 0.79 & 0.66 \\
LightningDiT~\citep{yao2025reconstruction}& 800 & 675M & 2.11 & 207.0 & 0.77 & 0.65 & 1.28 & 300.5 & 0.80 & 0.65 \\
RAE (DiT$^{\text{DH}})$~\citep{zheng2025diffusion} & 800 & 839M & 1.51 & 242.9 & 0.79 & 0.63 & 1.13 & 262.6 & 0.78 & \textbf{0.67}\\
\textbf{LightningDiT+IG (ours)} & 680 & 678M & \textbf{1.24} & 229.6 & 0.78 & 0.66 & \textbf{1.07} & 274.1 & 0.79 & 0.66 \\
\arrayrulecolor{black}\bottomrule
\end{tabular}
}
}
\vspace{-0.2cm}
\end{table*}
\section{Evaluation Metric}\label{f}
We strictly follow the setup and use the same reference batches of ADM \cite{dhariwal2021diffusion} for evaluation, following their official implementation.
We use NVIDIA A6000 pro GPUs or
4090Ti GPUs for evaluation and enable tf32 precision for faster generation.
In what follows, we explain the main concept of metrics that we used for the evaluation.

\begin{itemize}
\item \textbf{FID} \cite{heusel2017gans} measures the feature distance between the distributions of real and
generated images. It uses the Inception-v3 network \cite{szegedy2016rethinking} and computes distance
based on an assumption that both feature distributions are multivariate gaussian distributions.
\item \textbf{sFID} \cite{nash2021generating} proposes to compute FID with intermediate spatial features of the
Inception-v3 network to capture the generated images’ spatial distribution. 
\item \textbf{IS} \cite{salimans2016improved} also uses the Inception-v3 network but use logit for evaluation of the
metric. Specifically, it measures a KL-divergence between the original label distribution and the
distribution of logits after the softmax normalization.
\item \textbf{Precision and recall} \cite{kynkaanniemi2019improved} are based on their classic definitions: the
fraction of realistic images and the fraction of training data manifold covered by generated data.
\end{itemize}

As stated in~\citep{zheng2025diffusion}, using class-balanced sampling can lead to better results compared to random sampling because it more closely approximates the true label distribution of the training set.
As the FID value approaches a lower range, these subtle differences begin to have a greater impact.
The vast majority of VQ-based generation methods~\citep{tian2024visual,li2024autoregressive,ren2025beyond} have adopt the balanced sampling. 
To conduct a comprehensive and fair comparison with more baselines, we also adopt balanced sampling to conduct a comprehensive comparison of the current state-of-the-art VQ and Diffusion-based methods, as shown in Table \ref{tab:bc}.

\section{Computational Cost Comparison}\label{a}
We compare the computational efficiency of SiT-XL/2+REPA \cite{yu2024representation} and SiT-XL+IG under the same model scale in Table \ref{tab:flops}. IG introduces only a marginal increase in parameter count and FLOPs relative to SiT-XL, while maintaining neral identical latency.
Despite the minimal computational overhead, IG yields substantial improvements in generation quality, achieving a  relative reduction in FID, alongside an increase in IS. These results demonstrate that IG simultaneously improves generation quality
and computational efficiency, highlighting its effectiveness as a general-purpose enhancement for generative models.

\begin{table}[h!]
    \centering
    \caption{\textbf{Computational cost and performance comparison.} This table compares REPA and IG on ImageNet 256$\times$256, detailing model size, FLOPs, sampling steps, latency, and generation quality metrics. the proposed IG achieves substantially better sample quality with negligible increases in computational cost.}
    \scriptsize
    \setlength{\tabcolsep}{2pt}
    \resizebox{0.6\linewidth}{!}
    {
      \begin{tabular}{l|ccc|cc}
        \hline
        Method &\#Params & FLOPs$\downarrow$  &Latency (s)$\downarrow$  & FID$\downarrow$  & IS$\uparrow$ \\
        \hline
        SiT-XL/2 + REPA  &675 &114.46   & 6.18 &5.90 &157.8   \\
        \hline
        \multirow{2}{*}{SiT-XL/2 + IG}  &\multirow{2}{*}{{\shortstack[c]{678\\ (+0.44\%)}}} &\multirow{2}{*}{{\shortstack[c]{114.47\\ (+0.01\%)}}}  &\multirow{2}{*}{{\shortstack[c]{6.19\\ (+0.16\%)}}} &\multirow{2}{*}{{\shortstack[c]{1.75\\ (-70.34\%)}}}  &\multirow{2}{*}{{\shortstack[c]{228.6 \\ (+44.87\%)}}}  \\
        &&&&\\
        \hline
      \end{tabular}
    }
    \label{tab:flops}
\end{table}

\section{Comparison of Guidance Methods}\label{b}
\begin{wraptable}{r}{0.38\textwidth}
\vspace{-0.5cm}
	\small
 \renewcommand{\arraystretch}{1.}
	\renewcommand{\tabcolsep}{1pt}	
  \caption{Comparision with original Autoguidance on the EDM2-S model on ImageNet 64 dataset.}\label{oag}
\vspace{-0.3cm}
  \hspace{-.6cm}
    \begin{tabular}{c||c|c|c}
    \toprule
    Model&EDM2-S&EDM2-S+AutoGuidance&EDM2-S+IG   \\
    \midrule
    FID$\downarrow$& 1.58 & 1.01 & \textbf{0.99}  \\
    \bottomrule
  \end{tabular}
    \vspace{-0.5cm}
\end{wraptable}
To further illustrate the advantages of our proposed IG, we compared it with the original
CFG \cite{ho2022classifier} and Autoguidance \cite{karras2024guiding} methods.
For CFG method, we use the vanilla LightningDiT-XL/1 model which is trained for 800 epochs and employed its official CFG coefficients. 
For the Autoguidance method, since it is difficult to determine the optimal bad version model, we initialized a 12-layer LightningDiT model and trained it for 10 epochs. 
We roughly searched for the optimal coefficients of Autoguidance and determined it to be 1.15.
%
We also compare our IG with original Autoguidance on the vanilla EDM2-S~\cite{karras2024analyzing} model on ImageNet 64 dataset with the same training setting, which is shown in Table \ref{oag}.
As shown in Table \ref{tab:guidance}, our IG method outperforms both adding only CFG or only adding Autoguidance. 
Moreover, further combining CFG can achieve the current state-of-the-art in image generation, and has a significant improvement compared to other guidance methods.
\begin{table}[h!]
    \centering
    \caption{\textbf{Different guidance strategies comparison.} This table compares various sampling guidance methods. We compare the results of the standard CFG, Autoguidance, as well as our proposed IG and the combination of IG and CFG. Our proposed IG is more flexible compared to Autoguidance and can significantly enhance the results when combined with CFG as a plug-and-play approach.} 
    \scriptsize
    \setlength{\tabcolsep}{3pt}
    \resizebox{0.7\linewidth}{!}
    {
      \begin{tabular}{l|c|ccccc}
        \hline
        Method &Epochs&\textbf{FID}$\downarrow$& \textbf{sFID}$\downarrow$ & \textbf{IS}$\uparrow$ & \textbf{Prec.}$\uparrow$ & \textbf{Rec.}$\uparrow$  \\
        \hline
        LightningDiT-XL/1 + CFG~\citep{yao2025reconstruction} &800 &1.35 &4.15   & \textbf{295.3} &0.79 &0.65   \\
        LightningDiT-XL/1 + Autoguidance~\citep{karras2024guiding} &680  &1.81 &4.06  & 215.9 &0.79 &0.63   \\
        LightningDiT-XL/1 + IG (ours) & 680   &1.34 &\textbf{3.94}   &  229.3 &0.78 &0.65   \\
        LightningDiT-XL/1 + IG + CFG (ours)& 680 &\textbf{1.19} &4.11   & 269.0 &\textbf{0.79} & \textbf{0.66}   \\
        \hline
      \end{tabular}
    }
    \label{tab:guidance}
\end{table}
\section{512×512 ImageNet}\label{c}
To further validate IG's effectiveness, we conduct experiments at 512×512 resolution following REPA’s protocol \cite{yu2024representation} with Muon optimizer \cite{jordan2024muon}.
The RGB images are processed through the VAE \cite{rombach2022high} to yield 64×64×3 latents.
As demonstrated in Table, IG surpasses the performance of REPA trained for 200 epochs and SiT-XL/2 trained for 600 epochs in terms of FID at only 60 epochs, demonstrating its superior effectiveness.
\begin{figure*}[t]
\centering
\includegraphics[width=1\textwidth]{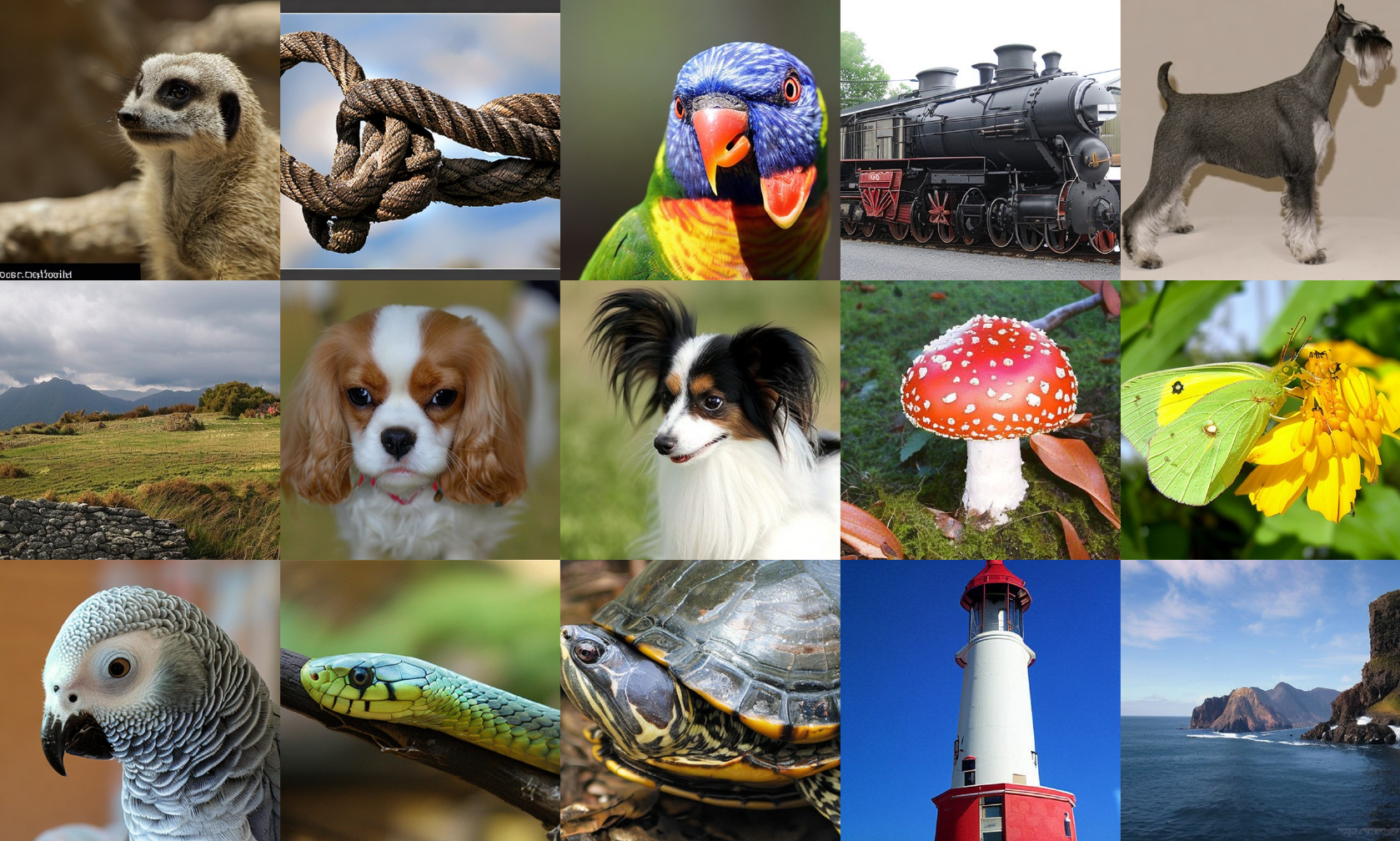}
\caption{\textbf{Samples on ImageNet 512×512} from SiT-XL/2 + IG (1.4) using CFG with $w = 1.8$.}
\vspace{-0.2cm}
\label{fig:visual}
\end{figure*}
\begin{table}[h!]
\centering\small
\caption{\textbf{Performance comparison} on ImageNet 512$\times$512. 
}
\begin{tabular}{l c c c c c c}
\toprule
{\pz\pz Model} & Epochs  &  {\pz FID$\downarrow$} & {sFID$\downarrow$} & {IS$\uparrow$} & {Pre.$\uparrow$} & Rec.$\uparrow$ \\
\arrayrulecolor{black}\midrule
\multicolumn{7}{l}{\emph{Pixel diffusion}\vspace{0.02in}} \\
\pz\pz {VDM$++$}                   & -   & 2.65 & -    & 278.1 & -    & - \\
\pz\pz {ADM-G, ADM-U}              & 400 & 2.85 & 5.86 & 221.7 & 0.84 & 0.53 \\
\pz\pz Simple diffusion (U-Net)    & 800 & 4.28 & -    & 171.0 & - & - \\
\pz\pz Simple diffusion (U-ViT, L) & 800 & 4.53 & -    & 205.3 & - & - \\
\arrayrulecolor{black!40}\midrule
\multicolumn{7}{l}{\emph{Latent diffusion, Transformer}\vspace{0.02in}} \\
\pz\pz MaskDiT                     & 800 & 2.50 & 5.10 & 256.3 & 0.83 & 0.56 \\ 
\arrayrulecolor{black!30}\cmidrule(lr){1-7}
\pz\pz {DiT-XL/2}                  & 600 & 3.04 & 5.02 & 240.8 & 0.84 & 0.54  \\
\arrayrulecolor{black!30}\cmidrule(lr){1-7}
\pz\pz {SiT-XL/2}                  & 600 & 2.62 & 4.18 & 252.2 & 0.84 & 0.57 \\
{\pz\pz{+ REPA}}         &\pz80 & {2.44} & 4.21 & 247.3 &  0.84 & 0.56   \\
{\pz\pz{+ REPA}}         & 100  &  {2.32} & 4.16 & 255.7 & 0.84 & 0.56  \\
{\pz\pz{+ REPA}}         & 200  & 2.08 & 4.19 & 274.6 & 0.83 & 0.58 \\
{\pz\pz{\textbf{+ IG}}}         & \textbf{60}  & \textbf{1.78} & \textbf{3.92} & \textbf{286.3} & 0.81 & \textbf{0.62} \\
\arrayrulecolor{black}\bottomrule
\end{tabular}
\label{tab:512}
\end{table}

\section{Hyperparameter and More Implementation Details}\label{d}
\paragraph{More implementation details.}
We implement our models based on the original DiT, SiT and LightningDiT implementation.
To speed up training and save GPU memory, we use mixed-precision (fp16) with 
gradient clipping and FlashAttention \cite{dao2023flashattention} operation for attention computation. 
We also pre-compute compressed latent vectors from raw pixels via stable diffusion VAE and VA-VAE and use these latent vectors.
We do not apply any data augmentation as discussed in MaskDiT and REPA.
For stable diffusion VAE, we use \texttt{stabilityai/sd-vae-ft-ema} for encoding images to latent vectors and decoding latent vectors to images.
For the output head, we use a Linear layer to map the feature to the output size.

\paragraph{The use of Muon}
In our large-scale experiment with the LightningDiT-XL/1 model, we adopt the Muon optimizer \cite{jordan2024muon}. We find that in the environment of PyTorch, when using the pre-trained self-supervised representation model (such as DINOv2-B \cite{oquab2023dinov2}) to align the encoder of the VAE, there will be unstable training issues in the early stage of training. 
After careful investigation and experimentation, we discover that replacing the AdamW optimizer with Muon could effectively alleviate such issue.
Compared to AdamW, the Muon optimizer has a certain acceleration convergence effect in the early stage of model training, but we observe that the gap between them gradually narrowed as the training time increased, reaching similar final convergence results. In the later stage of training, there is a small probability of encountering training instability issues again.
Resuming the training with full precision can solve this problem.
\paragraph{Computing resources.}
We use NVIDIA A6000 pro 96 GB GPUs for training largest model (LightningDiT-XL/1, SiT-XL/2); and uses NVIDIA 4090 24GB GPUs for training smaller model (L, B). When sampling, we use either NVIDIA A6000 pro 96 GB GPUs or NVIDIA RTX 4090 24GB GPUs to obtain the samples for evaluation.
\begin{table}[h!]
    \vspace{-1em}
    \centering
    \caption{\textbf{Defualt hyperparameter setup.} Unless other otherwise specified, we use these sets of hyperparameters for different models. In our ablation experiments, settings are also kept the same except those we point out in Table~\ref{tab:detailed_design}.} \label{tab:detailed_design}
    \vspace{0.3em}
    \makebox[\textwidth][c]{
    \resizebox{1\textwidth}{!}{
    \begin{tabular}{l c c c c c c}
        \toprule
         & SiT-B & SiT-L & SiT-XL & DiT-B & DiT-L & LightningDiT-XL \\
        \midrule
        \textbf{Architecture} \\
        Input dim. & 32$\times$32$\times$4 & 32$\times$32$\times$4 & 32$\times$32$\times$4 & 32$\times$32$\times$4 & 32$\times$32$\times$4 & 16$\times$16$\times$32\\
        Patch size & 2 & 2 & 2 & 2 & 2 & 2\\
        Num. layers & 12 & 24 & 28 & 12 & 24 & 28\\
        Hidden dim. & 768 & 1024 & 1152 & 768 & 1024 & 1152 \\
        Num. heads & 12 & 16 & 16 & 12 & 16 & 16\\ 
        \midrule
        \textbf{Optimization} \\
        Batch size & 256 & 256 & 256 & 256 & 256 & 1024\\ 
        Optimizer & AdamW & AdamW & AdamW & AdamW & AdamW & Muon \\
        lr & 0.0001 & 0.0001 &  0.0001 & 0.0001 & 0.0001 & 0.0002\\
        $(\beta_1, \beta_2)$ & (0.9, 0.999) & (0.9, 0.999) & (0.9, 0.999) & (0.9, 0.999) & (0.9, 0.999) & (0.9, 0.95)\\
        EMA decay & 0.9999  & 0.9999  & 0.9999  & 0.9999  & 0.9999  & 0.9995 \\
        \midrule
        \textbf{Interpolants or Denoising} \\
        $\alpha_t$ & $1-t$ & $1-t$ & $1-t$ & - & - & $1-t$ \\
        $\sigma_t$ & $t$ & $t$ & $t$ & - & - & $t$ \\
        $w_t$ & $\sigma_t$ & $\sigma_t$ & $\sigma_t$ & - & - &$\sigma_t$\\
        T & - & - & - & 1000 & 1000 & - \\
        Training objective & v-prediction & v-prediction & v-prediction & noise-prediction & noise-prediction & v-prediction\\
        Sampler & Euler-Maruyama & Euler-Maruyama & Euler-Maruyama & DDPM & DDPM & Heun \\
        Sampling steps & 250 & 250 & 250 & 250 & 250 & 125\\
        CFG Scale& - & - & 1.35 (if used) & - & - & 1.45 (if used) \\
        \midrule
        \textbf{IG} \\
        Auxiliary supervision position & 4  & 8 & 8  & 4  & 8 & 8 \\
         IG Scale& 2.3 & 2.3 & 1.4 & 2.3 & 2.3 & 1.4 \\
        \bottomrule
    \end{tabular}}}
    \vspace{-0.5em}
\end{table}

\section{Details of the 2D toy example}\label{e}
In this Section, we describe the construction of the 2D toy dataset used in the analysis of Section 4, as well as the associated model architecture, training setup, and sampling parameters.

\paragraph{Dataset.}
For each of the two classes $\mathbf{c}$, we model the fractal-like data distribution as a mixture of Gaussians
\mbox{$\mathcal{M}_\mathbf{c} = \big( \{\phi_i\}, \{\mathbf{\mu}_i\}, \{\mathbf{\Sigma}_i\} \big)$},
where $\phi_i$, $\mathbf{\mu}_i$, and $\mathbf{\Sigma}_i$ represent the weight, mean, and $2{\times}2$ covariance matrix of each component $i$, respectively.
This lets us calculate the ground truth scores and probability densities analytically and, consequently, to visualize them without making any additional assumptions.
The probability density for a given class is given by

\begin{equation}
p_{\text{data}}(\mathbf{x}\mid \mathbf{c})
= \sum_{i \in \mathcal{M}_c} \phi_i \,
\mathcal{N}(\mathbf{x}; \mu_i, \Sigma_i),
\quad\text{where}
\end{equation}

\begin{equation}
\mathcal{N}(\mathbf{x}; \mu, \Sigma)
= \frac{1}{\sqrt{(2\pi)^2 \det(\Sigma)}}
\exp\left(
-\frac{1}{2}(\mathbf{x}-\mu)^{\top} \Sigma^{-1} (\mathbf{x}-\mu)
\right).
\end{equation}

Applying heat diffusion to $p_\text{data}(\mathbf{x} | \mathbf{c})$, we obtain a sequence of increasingly smoothed densities $p(\mathbf{x} | \mathbf{c}; \sigma)$ parameterized by noise level $\sigma$:
\begin{equation}\label{eq:GT}
p(\mathbf{x}\mid \mathbf{c}; \sigma)
= \sum_{i \in \mathcal{M}_c} \phi_i \,
\mathcal{N}(\mathbf{x}; \mu_i, \Sigma_{i,\sigma}^*),
\quad\text{where } \Sigma_{i,\sigma}^* = \Sigma_i + \sigma^2 \mathbf{I}.
\end{equation}
The score function of $p(\mathbf{x} | \mathbf{c}; \sigma)$ is then given by
\begin{equation}
\nabla_{\mathbf{x}} \log p(\mathbf{x}\mid \mathbf{c}; \sigma)
=
\frac{
\sum_{i \in \mathcal{M}_c}
\phi_i \,
\mathcal{N}(\mathbf{x}; \mu_i, \Sigma_{i,\sigma}^*)
\, (\Sigma_{i,\sigma}^*)^{-1} (\mu_i - \mathbf{x})
}{
\sum_{i \in \mathcal{M}_c}
\phi_i \,
\mathcal{N}(\mathbf{x}; \mu_i, \Sigma_{i,\sigma}^*)
}
.
\end{equation}

We construct $\mathcal{M}_\mathbf{c}$ to represent a thin tree-like structure by starting with one main ``branch'' and recursively subdividing it into smaller ones.
Each branch is represented by 8 anisotropic Gaussian components and the subdivision is performed 6 times, decaying $\phi$ after each subdivision and slightly randomizing the lengths and orientations of the two resulting sub-branches.
This yields \mbox{$127 {\times} 8 = 1016$} components per class and $1016 {\times} 2 = 2032$ components in total.
We define the coordinate system so that the mean and standard deviation of $p_\text{data}$, marginalized over $\mathbf{c}$,
are equal to~$0$ and $\sigma_\text{data} = 0.5$ along each axis, respectively, matching the recommendations by Karras~et~al.~\cite{Karras2022elucidating}.

\paragraph{Models.}
We implement the denoiser models $D$ as simple multi-layer perceptrons, utilizing the magnitude-preserving design principles from EDM2~\cite{karras2024analyzing}.
Following Autoguidance \cite{karras2024guiding}, we design the model interface so that for a given noisy sample,
each model outputs a single scalar representing the logarithm of the corresponding unnormalized probability density, as opposed to directly outputting the denoised sample or the score vector.
Concretely, let us denote the output of a given model by $G_\theta(\mathbf{x}; \sigma, \mathbf{c})$.
The corresponding normalized probability density is then given by
\begin{equation}
p_{\theta}(\mathbf{x}\mid \mathbf{c}; \sigma)
=
\exp\!\big( G_{\theta}(\mathbf{x}; \sigma, \mathbf{c}) \big)
\;\Big/\;
\int \exp\!\big( G_{\theta}(\mathbf{x}; \sigma, \mathbf{c}) \big)\, d\mathbf{x}.
\end{equation}
By virtue of defining $G_\theta$ this way, we can derive the score vector, and by extension, the denoised sample, from $G_\theta$ through automatic differentiation:
\begin{equation}\label{eq:EBM}
\nabla_x \log p_{\theta}(x | c; \sigma) = \nabla_x G_{\theta}(x; \sigma, c)
\end{equation}

\begin{equation}
D_{\theta}(x; \sigma, c) = x + \sigma^2 \nabla_x G_{\theta}(x; \sigma, c).
\end{equation}
Besides Equation~\ref{eq:EBM}, 
according to the description of Autoguidance,
trying out the alternative formulations where the model outputs the score vector or the denoised sample directly is qualitatively more or less identical.

To connect the above definition of $G_\theta$ to the raw network layers, we apply preconditioning using the same general principles as in EDM~\cite{Karras2022edm}.
Denoting the function represented by the raw network layers as $F_{\theta}$, we define $G_\theta$ as
\begin{equation}\label{eq:precond}
G_{\theta}(\mathbf{x}; \sigma, c) = -\frac{1}{2}\|\mathbf{x}\|_2^2 - \frac{g_{\theta}}{\sigma n} \sum_{i=1}^{n} F_{\theta, i}\left(\mathbf{x}^{*}; \frac{1}{4}\log \sigma, c\right)^2, \quad \text{where} \quad \mathbf{x}^{*} = \frac{\mathbf{x}}{\sqrt{\sigma^2 + \sigma_{\text{data}}^2}}
\end{equation}
and the sum is taken over the $n$ output features of $F_\theta$.
We scale the output of $F_\theta$ by a learned scaling factor $g_\theta$ that we initialize to zero.
The goal of Equation~\ref{eq:precond} is to satisfy the following three requirements:
\begin{itemize}
\item The input of $F_\theta$ should have zero mean and unit magnitude. This is achieved through the division by $\sqrt{\sigma^2 + \sigma_\text{data}^2}$.
\item After initialization, $G_\theta$ should represent the best possible first-order approximation of the correct solution. This is achieved through the $-\frac{1}{2}\|\mathbf{x}^{*}\|_2^2$ term, as well as the fact that $g_\theta = 0$ after initialization.
\item After training, $\sqrt{g_\theta} \cdot F_\theta$ should have approximately unit magnitude. This is achieved through the division by $\sigma n$.
\end{itemize}

In practice, we use an MLP with one input layer and four hidden layers, interspersed with SiLU~\cite{hendrycks2016gaussian} activation functions and implemented using the magnitude-preserving primitives from EDM2~\cite{Karras2024edm2}.
Additionally, we defined an output layer after the first hidden layer to produce a weaker version of the output.
The input is a 4-dimensional vector $\big[ \mathbf{x}^\ast_x; \mathbf{x}^\ast_y; \tfrac{1}{4} \log \sigma; 1 \big]$ and the output of each hidden layer has $n$ features, where $n = 64$.

\paragraph{Training.}

Given that we have the exact score function of the ground truth distribution readily available (Equation~\ref{eq:GT}), we train the models using exact score matching~\cite{hyvarinen2005estimation} for simplicity and increased robustness.
We thus define the loss function and the additional supervision loss as
\begin{equation}
\mathcal{L}(\theta_i) = \mathbb{E}_{\sigma \sim p_{\text{train}}, \mathbf{x} \sim p(\mathbf{x}; \sigma)} \sigma^2 \left\|\nabla_{\mathbf{x}} \log p_{\theta_i}(\mathbf{x}; \sigma) - \nabla_{\mathbf{x}} \log p(\mathbf{x}; \sigma)\right\|_2^2,
\end{equation}

\begin{equation}
\mathcal{L}(\theta_d) = \mathbb{E}_{\sigma \sim p_{\text{train}}, \mathbf{x} \sim p(\mathbf{x}; \sigma)} \sigma^2 \left\|\nabla_{\mathbf{x}} \log p_{\theta_d}(\mathbf{x}; \sigma) - \nabla_{\mathbf{x}} \log p(\mathbf{x}; \sigma)\right\|_2^2,
\end{equation}
where $\sigma \sim p_\text{train}$ is realized as $\log(\sigma) \sim \mathcal{N}(P_\text{mean}, P_\text{std})$ \cite{Karras2022elucidating} and $\theta_i$ and $\theta_d$ represent the intermediate layer and the deep layer network respectively.
In particular, for the training acceleration method depicted in Figure 4, our loss function is
\begin{equation}
\mathcal{L}(\theta_i) = \mathbb{E}_{\sigma \sim p_{\text{train}}, \mathbf{x} \sim p(\mathbf{x}; \sigma)} \sigma^2 \left\|\nabla_{\mathbf{x}} \log p_{\theta_i}(\mathbf{x}; \sigma) - \nabla_{\mathbf{x}} \log p(\mathbf{x}; \sigma)\right\|_2^2,
\end{equation}
\begin{equation}
\mathcal{L}(\theta_d) = \mathbb{E}_{\sigma \sim p_{\text{train}}, \mathbf{x} \sim p(\mathbf{x}; \sigma)} \sigma^2 \left\|\nabla_{\mathbf{x}} \log p_{\theta_d}(\mathbf{x}; \sigma) - (\nabla_{\mathbf{x}} \log p(\mathbf{x}; \sigma)+0.5*(\nabla_{\mathbf{x}} \log p_{\theta_d}(\mathbf{x}; \sigma)-\nabla_{\mathbf{x}} \log p_{\theta_i}(\mathbf{x}; \sigma)))\right\|_2^2,
\end{equation}
The final training loss is $\mathcal{L} = \mathcal{L}(\theta_d) + 0.5*\mathcal{L}(\theta_i)$
The more commonly used denoising score matching do not show any significant differences in the model behavior or training dynamics.

We train these models for 4096 iterations using a batch size of 4096 samples. We set
  $P_\text{mean} \!= -2.3$ and $P_\text{std} \!= 1.5$,
  and use $\alpha_\text{ref} / \sqrt{\max(t / t_\text{ref}, 1)}$ learning rate decay schedule with $\alpha_\text{ref} \!= 0.01$ and $t_\text{ref} \!= 512$ iterations,
  along with a power function EMA profile~\cite{Karras2024edm2} with $\sigma_\text{rel} \!= 0.010$.
Overall, the setup is robust with respect to the hyperparameters;
the phenomena illustrated in Figures 3 and 4 remain unchanged across a wide range of parameter choices.
\paragraph{Sampling.}
We use the standard EDM sampler~\cite{Karras2022elucidating} with \mbox{$N = 32$} Heun steps (\mbox{$\text{NFE} = 63$}), \mbox{$\sigma_\text{min} \!= 0.002$}, \mbox{$\sigma_\text{max} \!= 5$}, and \mbox{$\rho = 7$}.
We chose the values of $N$ and $\sigma_\text{max}$ to be much higher than what is actually needed for this dataset in order to avoid potential discretization errors from affecting our conclusions.
In Figure 3, we set \mbox{$w = 2.5$} for CFG, \mbox{$w = 2$} for autoguidance and \mbox{$w = 2$} for internal guidance (IG), and \mbox{$w_1 = 1$} and \mbox{$w_2 = 1.5$} for IG + CFG.
In Figure 4, we set \mbox{$w = 2$} for internal guidance.

\section{More Discussion on Related Work}\label{g}
\paragraph{Denoising transformers.}
Many recent works have tried to use transformer backbones for diffusion or flow-based model training.
First, several works like U-ViT \cite{bao2023all}, MDT \cite{gao2023mdtv2}, and DiffiT \cite{hatamizadeh2024diffit} how transformer-based backbones with skip connections can be an effective backbone for training diffusion models.
Intriguingly, DiT and SiT show a pure transformer architecture can be a scalable architecture for training diffusion-based models.
Based on these improvements, Stable diffusion 3 \cite{esser2024scaling}, FLUX 1 \cite{labs2025flux1kontextflowmatching} and Nano Banana \cite{NanoBanana} show pure transformers can be scaled up for challenging text-to-image generation, and Sora, CogvideoX \cite{yang2024cogvideox} and Wan \cite{wan2025wan} demonstrate their success in text-to-video generation. 
Our work analyzes and improves the training of Diffusion Transformer architecture based on a simple auxiliary supervision to the early layers.
\paragraph{Sampling guidance in diffusion models.}
Achieving better generation results over diffusion models remains challenging yet essential.
Early approaches, such as classifier guidance \cite{dhariwal2021diffusion}, introduce control by incorporating classifier gradients into the
sampling process. However, this method requires separately
trained classifiers, making it less flexible and computationally demanding. To overcome these limitations, classifier free guidance (CFG) \cite{ho2022classifier} is proposed, enabling guidance
without the need for an external classifier. Instead, CFG
trains conditional and unconditional models simultaneously
and interpolates between their outputs during sampling.
Subsequently, several studies \cite{chang2023muse, gao2023masked, kynkaanniemi2024applying, sadat2023cads, wang2024analysis, chung2024cfg++, fan2025cfg} have reduced the downsides of CFG by making the guidance weight noise level-dependent.
Another research direction separates the class guidance part from the CFG and only uses a weaker version of the current model for guidance \cite{hong2023improving, hong2024smoothed, karras2024guiding, hyung2025spatiotemporal, chen2025s}, which also bears conceptual similarity to contrastive decoding \cite{li2023contrastive} used in large language models to reduce the repetitiveness of generations.
Our approach does not require any additional sampling steps. It can be applied to any deep diffusion Transformer network in a plug-and-play manner while achieving better generation results.
\paragraph{Efficient training of generative models.}
The issue of how to accelerate the convergence of generative models during the training process has also received significant attention.
Previous studies have explored architectural optimization \cite{peebles2023scalable, tian2025dic, krause2025tread}, improved flow-based theories \cite{lipman2022flow,fischer2023boosting}, optimized data \cite{tong2023improving,zhang2017mixup} to accelerate convergence.
Meanwhile, there have been several approaches in generative adversarial network \cite{goodfellow2014generative} that try to accelerate training with
better convergence using pretrained visual encoder \cite{sauer2021projected, kumari2022ensembling, sauer2022stylegan, kang2023scaling}.
Another line of work tries to exploit the pretrained visual encoders for improving diffusion model training from scratch \cite{pernias2023wurstchen, li2024return}, usually by training two diffusion models where one model generates the pretrained representations and the other model generates the target data conditioned on the generated representation.
More recently, representation learning has been introduced into generative model training to further enhance training efficiency \cite{NEURIPS2024_e304d374,yu2024representation, leng2025repa, wu2025representation,yao2025reconstruction,zheng2025diffusion}.
Our method not only alleviate the problem of gradient vanishing during training, but also can be directly applied in the sampling stage to achieve better generation results on models with less training consumption. 
\section{More Qualitative Results}\label{h}
Below we show some uncurated generation results on ImageNet 256×256 from the SiT-XL+IG (1.8).
We use classifier-free guidance with w = 2.5.
\begin{figure}[h]
    \centering
    \includegraphics[width=1\linewidth]{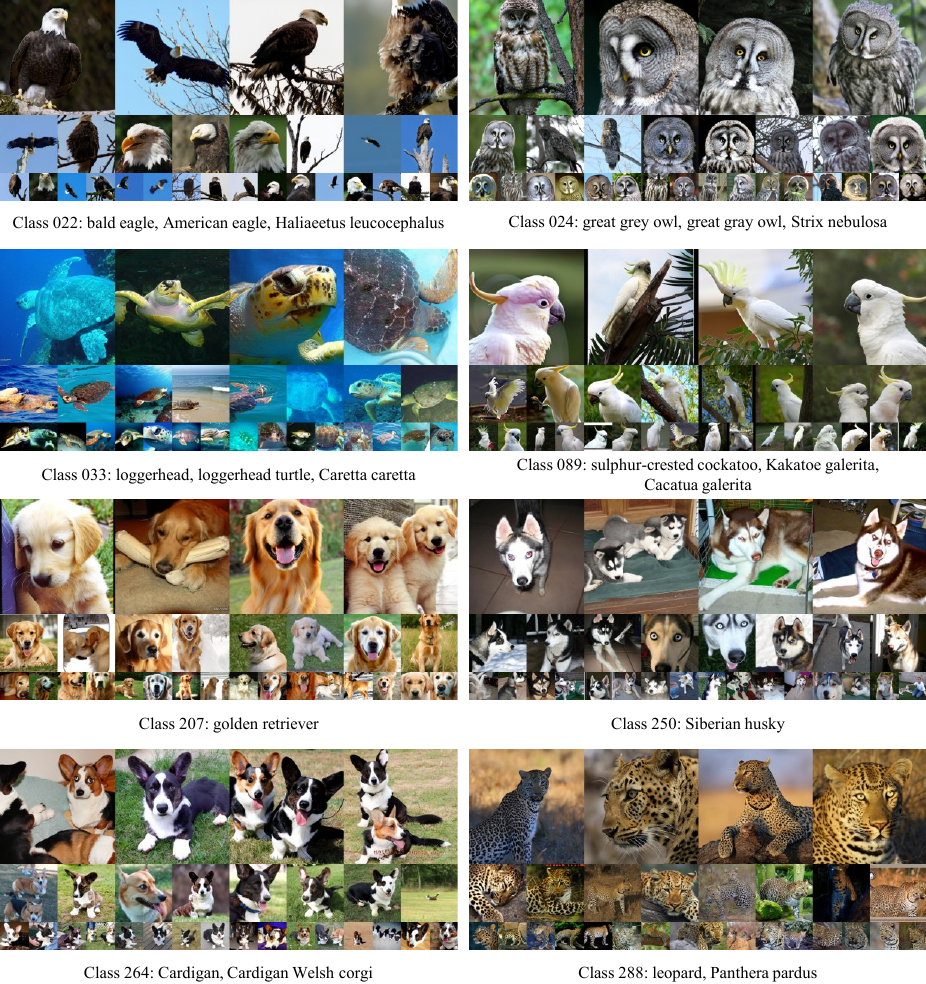}
    \caption
    {
        Uncurated visualization results of LightningDiT-XL/1 + IG (1.8) use CFG with $w = 2.5$.
    }
    \label{fig:app_com5}
    \vspace{-20pt}
\end{figure}
\begin{figure}
    \centering
    \includegraphics[width=1\linewidth]{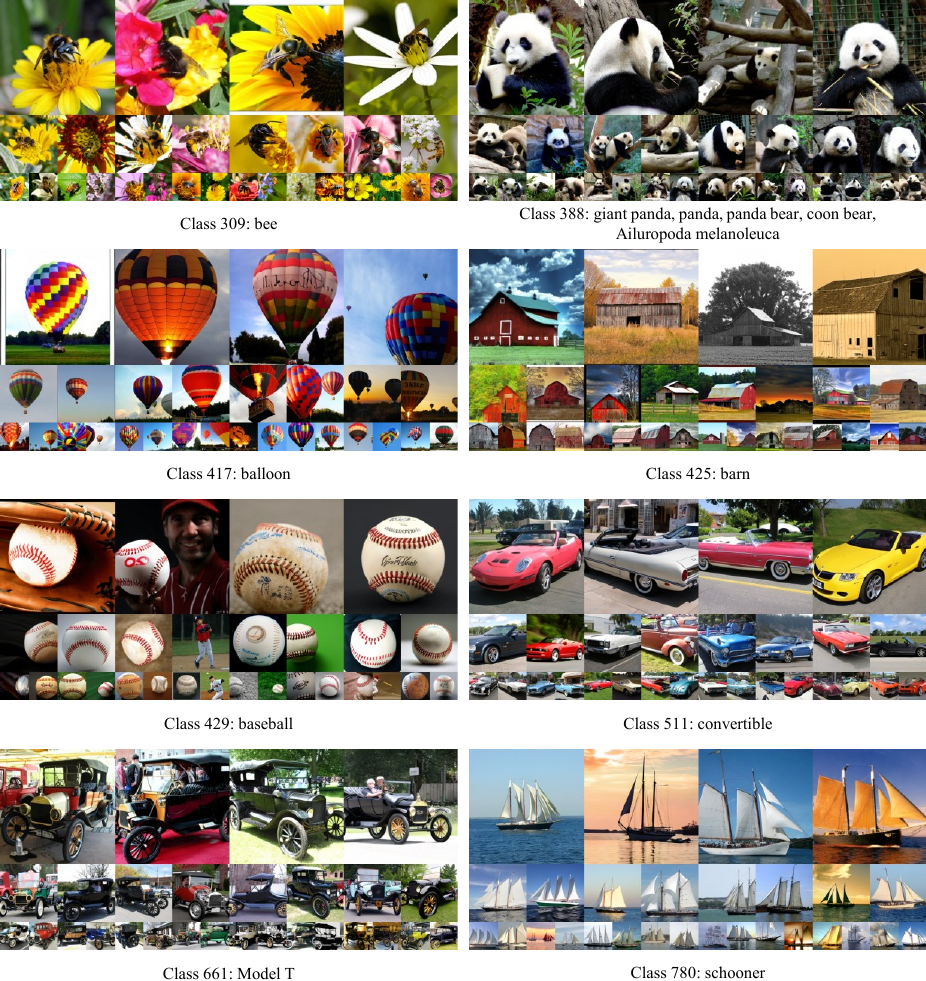}
    \caption
    {
         Uncurated visualization results of LightningDiT-XL/1 + IG (1.8) use CFG with $w = 2.5$.
    }
    \label{fig:app_com5}
    \vspace{-20pt}
\end{figure}
\begin{figure}
    \centering
    \includegraphics[width=1\linewidth]{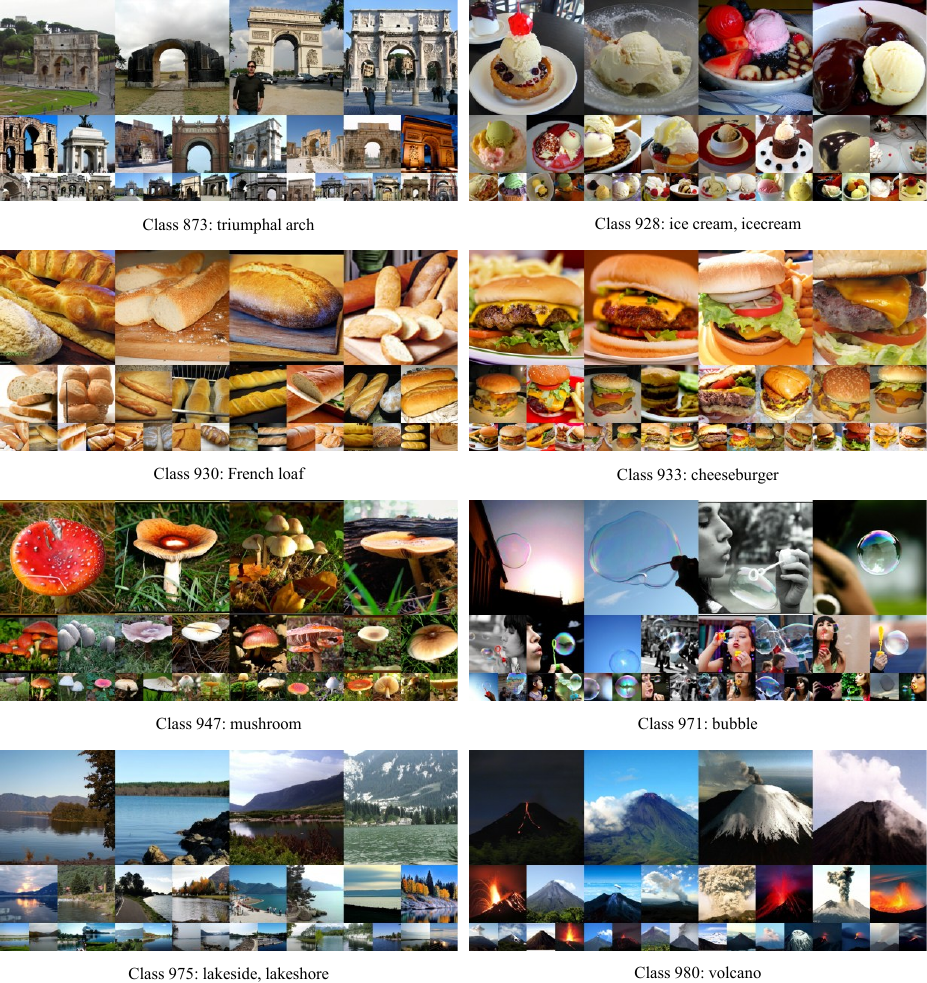}
    \caption
    {
        Uncurated visualization results of LightningDiT-XL/1 + IG (1.8) use CFG with $w = 2.5$.
    }
    \label{fig:app_com5}
    \vspace{-20pt}
\end{figure}

\end{document}